\documentclass[12pt,twocolumn,twoside]{IEEEtran}
\usepackage{cite}
\usepackage{amsmath}
\usepackage{amssymb}
\usepackage{fontenc}
\usepackage{graphicx}
\usepackage{epstopdf}
\usepackage{color}

\newtheorem{remark}{Remark}
\usepackage{algorithm}
\usepackage{algorithmic}
\usepackage[caption=false,font=footnotesize]{subfig}
\usepackage{stfloats}
\usepackage{url}
\usepackage{booktabs}
\usepackage{balance}
\usepackage{float}

\usepackage{xcolor}

\begin{document}

\title{Distributed Reinforcement Learning for Privacy-Preserving Dynamic Edge Caching}

\author{Shengheng~Liu,~\IEEEmembership{Member,~IEEE},
        Chong~Zheng,~\IEEEmembership{Student~Member,~IEEE},
        Yongming~Huang,~\IEEEmembership{Senior Member,~IEEE},
        and~Tony~Q.~S.~Quek,~\IEEEmembership{Fellow,~IEEE}

\thanks{Manuscript received February 21, 2021; revised November 2, 2021; accepted XXX XX, XXXX. Date of publication XXX XX, XXXX; date of current version XXX XX, XXXX. This work was supported in part by the National Natural Science Foundation of China under Grant Nos. 62001103 and the National Key R\&D Program of China under Grant No. 2020YFB1806600. Part of this work has been accepted for presentation at the IEEE Global Communications Conference (GLOBECOM): Machine Learning for Communications Symposium, Madrid, Spain, December 2021 \cite{GlobeCom21}. (Corresponding author: Y.~Huang.)}

\thanks{S.~Liu, C.~Zheng, and Y.~Huang are with the School of Information Science and Engineering, Southeast University, Nanjing 210096, China, and also with the Purple Mountain Laboratories, Nanjing 211111, China (e-mail: \{s.liu; czheng; huangym\}@seu.edu.cn).}
\thanks{T.~Q.~S.~Quek is with the Information System Technology and Design Pillar, Singapore University of Technology and Design, Singapore 487372 (e-mail: tonyquek@sutd.edu.sg).}
}

\markboth{IEEE JOURNAL ON SELECTED AREAS IN COMMUNICATIONS,~Vol.~XX, No.~X, XXX~2022}%
{Liu \MakeLowercase{\textit{et al.}}: Distributed Reinforcement Learning for Privacy-Preserving Dynamic Edge Caching}

\maketitle
\begin{abstract}
Mobile edge computing (MEC) is a prominent computing paradigm which expands the application fields of wireless communication. Due to the limitation of the capacities of user equipments and MEC servers, edge caching (EC) optimization is crucial to the effective utilization of the caching resources in MEC-enabled wireless networks. However, the dynamics and complexities of content popularities over space and time as well as the privacy preservation of users pose significant challenges to EC optimization. In this paper, a privacy-preserving distributed deep deterministic policy gradient (P2D3PG) algorithm is proposed to maximize the cache hit rates of devices in the MEC networks. Specifically, we consider the fact that content popularities are dynamic, complicated and unobservable, and formulate the maximization of cache hit rates on devices as distributed problems under the constraints of privacy preservation. In particular, we convert the distributed optimizations into distributed model-free Markov decision process problems and then introduce a privacy-preserving federated learning method for popularity prediction. Subsequently, a P2D3PG algorithm is developed based on distributed reinforcement learning to solve the distributed problems. Simulation results demonstrate the superiority of the proposed approach in improving EC hit rate over the baseline methods while preserving user privacy.
\end{abstract}


\bigskip

\section*{{Nomenclature}}

{For ease of reading, at the top of next page, a nomenclature of notations that will be later used within the body of this paper is given.}

\begin{table*}[!t]
  {\centering
  \begin{tabular}{cl}
              $t$ &  Index of time slot. \\
          ${\cal I} = \{ 1,2, \cdots ,I\}$ &  Set of UE labels. \\
      ${{\cal F}} = \{ {{F}_1},{{ F}_2}, \cdots ,{{F}_N}\}$ &  Set of all contents. \\
      ${M_0}$, ${M_i}$  & Storage capacities of the MEC server and {UE}-$i$, respectively. \\
           $\mathcal{C}_{0}(t)$, $\mathcal{C}_{i}(t)$ &  Content sets respectively cached in the MEC server and {UE}-$i$ at time $t$. \\
      ${F^i}(t)$ & Content request generated by {UE}-$i$ at time $t$. \\
      $\lambda_{i}\left(t\right)$ & Content request's arrival rate associated with {UE}-$i$ at time $t$.\\
      ${{\bf{P}}^{\rm{G}}}(t) = [P_n^{\rm{G}}(t)]_{n = 1}^N$, ${\bf P}^{i}\left(\alpha^{i}\left(t\right),t\right)=\left\{ P_{n}^{i}\left(\alpha^{i}\left(t\right),t\right)\right\} _{n=1}^{N}$ &  Content popularities of the MEC server and {UE}-$i$ at time $t$, respectively.  \\
      $\alpha^{i}\left(t\right)$&  Distribution parameter of {UE}-$i$'s popularity at time $t$. \\
      ${{\cal G}_i} = \{ \alpha _{g_i}^i|g_i = 1,2, \cdots ,{G_i}\}$ & Parameters set that $\alpha^{i}\left(t\right)$ evolves over time.\\
      ${\bf{R}}^{\rm{G}}(t) = \left\{{{ F}^i}(t)\right\}_{i = 1}^I$         &  Request information received by the MEC server at time $t$. \\
      $\mathcal{C}_{0}^{\textrm{u}}(t)$, $\mathcal{C}_{i}^{\textrm{u}}(t)$  & Sets of requested but not cached contents at MEC server and {UE}-$i$, respectively. \\
      $H_{0}\left(t\right)$, $H_{i}\left(t\right)$ & Realtime cache hit rates at MEC server and {UE}-$i$ sides, respectively.  \\
      $H_{i}^{\textrm{savg}}\left(t\right)$             &  Sliding average of $H_{i}\left(t\right)$ over a period of time $T_{\textrm{h}}$. \\
      $\mathbf{a}_{0}\left(t\right)\!=\!\left(\mathbf{a}_{0}^{+}\left(t\right)\!,\mathbf{a}_{0}^{-}\left(t\right)\right)$, $\mathbf{a}_{i}\left(t\right)\!=\!\left(a_{i}^{+}\left(t\right)\!,\mathbf{a}_{i}^{-}\left(t\right)\right)$  &  Dynamic caching actions of the MEC server and {UE}-$i$ at time $t$, respectively.\\
      $\mathcal{A}_{0}$, $\mathcal{A}_{i}$ &  Collections of $\mathbf{a}_{0}\left(t\right)$ and $\mathbf{a}_{i}\left(t\right)$, respectively, in each time slot $t$. \\
      $\mathcal{S}_{0} = \left\{\mathbf{s}_{0}\left(t\right) \mid t = 0,1,\cdots\right\}$, $\mathcal{\widetilde{S}}_{0}=\left\{\mathbf{\widetilde{s}}_{0}\left(t\right)\mid t=0,1,\cdots\right\}$                 & State space and its renewed version at the MEC server side.  \\
      $\mathcal{S}_{i} = \left\{\mathbf{s}_{i}\left(t\right) \mid t = 0,1,\cdots\right\}$, $\mathcal{\widetilde{S}}_{i}=\left\{\mathbf{\widetilde{s}}_{i}\left(t\right)\mid t=0,1,\cdots\right\}$                 &  State space and its renewed version at the local {UE}-$i$ side. \\
      $r_{0}\left(t\right)$, $r_{i}\left(t\right)$  &  Cumulative reward starting from time $t$ at global and {UE}-$i$ sides, respectively.\\
      ${{\bf{R}}^i}(t) = [{{ F}^i}(t - H), \cdots ,{{ F}^i}(t)]$                 &  Extractor of {UE}-$i$'s historical request information.\\
      ${{\Theta ^{\rm{G}}}}$, ${{\Theta ^{\rm{i}}}}$ & Parameters sets of  global and local popularity prediction models. \\
      $\Theta^{\textrm{A}}$, $\overline{\Theta}^{\textrm{A}}$  &  Trainable parameter sets of the online and target actor networks. \\
      $\Theta^{\textrm{C}}$, $\overline{\Theta}^{\textrm{C}}$          &  Trainable parameter sets of the online and target critic networks. \\
      $\pi_{0}$, $\pi_{i}$   &  Dynamic caching policies of the MEC server and {UE}-$i$. \\
      $V^{\pi_{0}}\left(\cdot\right)$, $V^{\pi_{i}}\left(\cdot\right)$              &  Value function under policies $\pi_{0}$ and $\pi_{i}$, respectively.\\
      $\pi^{\Theta^{\textrm{A}}}\left(\cdot\right)$, $Q\left(\left.\cdot\right|\Theta^{\textrm{C}}\right)$  &  Parameterized online actor and target networks.\\
      $\mathbf{n_{0}}\left(t\right)$         &  Gaussian noise vector at time $t$.\\
      $\Omega$   &  Replay buff for training at the MEC server. \\
      $L\!\left(\!\Theta^{\textrm{C}}\!\right)$  &  Training loss function of the online critic network. \\
      $J_{\beta}\left(\pi\right)$                 &  Performance objective function for the current policy evaluation. \\
      $\chi$ &  Discount factor in cumulative reward. \\
      $\varPsi$  &  Total episodes of training. \\
      $\varphi$&  Step interval between online/target networks in parameter clone.\\
      $\nu$  &  Soft-update coefficient.\\
\end{tabular}}
\end{table*}

\section{Introduction}
\label{sec1}

\IEEEPARstart{W}{ith} the rapid proliferation of advanced wireless applications such as virtual reality and Internet of vehicles (IoV), the demand of delay-sensitive and computation-intensive data services in mobile networks has been soaring at an unprecedented pace \cite{HuF20, Duan20, Abde19}. Along with the advent of beyond fifth-generation (B5G) communications, the increasing speed of this demand will achieve a further leap and pose significant challenges for the computing and caching capabilities of wireless communication systems. A promising network paradigm to tackle this challenge is mobile edge computing (MEC) \cite{Fara20, DuJ20}. By equipping the processing servers with the edge nodes (ENs), i.e., WiFi access point or micro base station, MEC framework provides cloud-computing/caching capabilities within the radio access network in close proximity to terminal devices, thereby greatly reducing the service latency as well as mitigating the surging cache and computation burden of the data centers \cite{Xiong20,Miao18,Zhao20}. Furthermore, edge caching (EC) as one of the key techniques in MEC networks can sufficiently exploit the caching resources in edge networks to promote caching efficiency of the ENs and user equipments (UEs) \cite{Wang20} and further reduce the latency.

Recently, the explorations of optimal caching placement policies of EC from the perspective of the relationships among the contents, the ENs and the cloud center have been investigated in many works, i.e., \cite{cui16, Niko16, LiQ20}. In \cite{cui16}, the authors consider the analysis and optimization of EC and multicasting in a large-scale MEC-enabled wireless network. On the basis of file combinations, an iterative numerical algorithm is proposed in \cite{cui16} to maximize the successful transmission probability and obtain the local optimal caching and multicasting design. By leveraging social links between clients and ENs, cooperative cache placement schemes are developed to reduce client bandwidth overheads in \cite{Niko16}. Furthermore, the cooperation between ENs and cloud center is also studied in Li et al. \cite{LiQ20}. The authors in \cite{LiQ20} proposed a capacity-aware EC framework and formulated the average-download-time (ADT) minimization problem as a multi-class processor queuing process by allowing cooperation between ENs and cloud center. However, the mentioned works assumed that the content popularity is constant during the service and is known a priori, which is impractical. Generally, content popularity is time-invariant and unavailable in advance regardless of the caching policy used \cite{Jiang19}.

To consider time-varying content popularities, the complicated, subjective and dynamic preferences of users pose significant challenges to the effective design and optimization of the EC policies. To this end, dynamic caching replacement scheme which continuously updates the cache under certain replacement policies during the services is investigated to address these challenges \cite{Azimi18, Liu20}. The authors in \cite{Azimi18} focus on the scenario where the set of popular content is time-varying, hence they investigate the online replenishment of the ENs caches along with the delivery of the requested files. To minimize the long-term normalized delivery time, online EC and delivery schemes as well as the reactive and proactive online caching schemes are proposed \cite{Azimi18}. Liu et al. \cite{Liu20} leverages the estimation of popularity to improve the dynamic caching performance. Specifically, an online Bayesian clustering caching algorithm is introduced for the cache provider to autonomously learn the users' interactive cache hit data in a collaborative way while maintaining sustainable scalability. Nevertheless, the popularity of each cluster has to be {\em a priori} given in \cite{Liu20}, which is still challenging in practice.

On the other hand, privacy preservation in privacy-sensitive applications tightens the interactions among UES and servers in MEC systems to enhance user and data security. To ensure a secure EC in vehicle-to-vehicle based MEC network, Dai et al. \cite{DaiY20} propose a blockchain empowered distributed content caching framework where the content caching is performed in vehicles and the base stations (BSs) do not execute the content caching but just maintain permissioned blockchain to ensure an secure content caching in vehicles. However, the proposed blockchain-based EC scheme in \cite{DaiY20} sacrifices the cache capabilities of the BSs, which are far more than that of vehicles. Moreover, the time-varying characteristic of the content popularity is not considered in \cite{DaiY20}. In \cite{XuQ19}, the authors explore the privacy preservation in EC from the perspective of game theory, and propose a game theoretical secure caching scheme to guarantee the integrity of cached contents while preserving the privacy of users. It can be observed that the EC problem considering in \cite{XuQ19} is still a static caching problem where the cached content is locally encrypted on UEs to prevent leakage. The MEC server just cache the corresponding cryptograph for restoring original content, which leads to the same waste of cache resources as the scheme proposed in \cite{DaiY20}. Recently, machine learning (ML) has shown potential usefulness in privacy-preserving MEC systems \cite{Xiao18, XuQ19}. In \cite{YuZ20}, the authors propose a mobility-aware proactive caching scheme based on FL to dynamically update cached contents in the MEC servers according to the mobility and position information of vehicles. However, the caching scheme proposed in \cite{YuZ20} centrally caches contents in the MEC servers and ignores the abundant cache resources of the terminal devices.

In this paper, we present a privacy-preserving distributed deep deterministic policy gradient (P2D3PG) algorithm to solve the distributed cache hit rate maximization problems under the consideration of time-varying and unobservable content popularities as well as the constraints of user privacy preservation. Specifically, our contributions are summarized as follows:

\begin{itemize}
	\item  We formulate a distributed optimization problem to maximize the cache hit rate of all the cache entities in the MEC-enabled system and design a dynamic caching replacement mechanism to enhance the personalized utilization of the cache resources in the system.
		
	\item With the constraints of privacy preservation and dynamic content popularities, we convert the distributed optimization problem into a distributed model-free Markov decision process (MDP) problem and further introduces a privacy-preserving FL method to predict the distributed popularities.

	\item A P2D3PG algorithm is developed to maximize the EC hit rate of devices in the system in a distributed way without any privacy leakage. The P2D3PG algorithm addresses the challenges in extending the centralized deep deterministic policy gradient method to a distributed manner. The performance advantages in terms of the cache hit rate are also presented in the numerical results.
\end{itemize}

The remainder of this paper is organized as follows. The system model is presented in Section II. Then, Section III introduces the problem formulation and analysis. In Section IV, the P2D3PG algorithm is presented with details. In Section V, simulation results are discussed. Finally, conclusions are drawn in Section VI.

\section{SYSTEM MODEL}
\label{sec2}

\begin{figure}[!hpbt]
\centering
\includegraphics[width=0.48\textwidth]{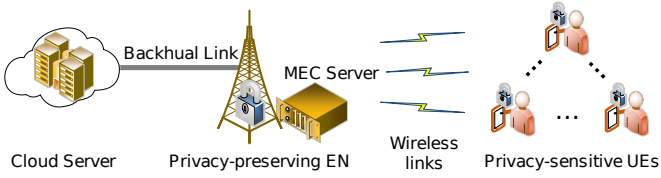}
\caption{Hierarchical architecture of the privacy-preserving MEC system under investigation.}
\label{fig1}
\end{figure}

In the following, we investigate the optimizations of EC policy in the privacy-preserving MEC system. Fig. \ref{fig1} illustrates the wireless service scenario in a privacy-preserving MEC network with $I$ privacy-sensitive UEs and one privacy-preserving EN, where the MEC server and all the UEs have certain computing and caching capabilities. For {UE}-$i$ at time $t$, once a content is requested but uncached locally, {UE}-$i$ will upload request information to access this uncached content from the MEC server. Limited by the caching capability, the MEC server also occasionally access to the cloud through the backhual link for absent contents if necessary. Due to our privacy-preserving mechanism, each privacy-sensitive UEs will protect its database of historical requests from snooping by outsiders. Furthermore, the privacy-preserving EN has no permission to retain any historical information of any UEs, and the current requests information from UEs at time $t$ must be immediately deleted from the MEC server once the contents have been scheduled.

\subsection{Service Process}\label{ser_pro}

Let ${{\cal F}} = \{ {{F}_1},{{ F}_2}, \cdots ,{{F}_N}\}$ denote the set of all contents and all these contents can be accessed from the cloud. We consider that the caching entities in the MEC server and each {UE}-$i$ with limited storage capacities of ${M_0}$ and ${M_i}$ contents respectively, where $\forall i \in {\cal I} = \{ 1,2, \cdots ,I\}$ is the set of {UE} labels. Without loss of generality, we assume that $M_{i}\ll M_{0}<N$. At time $t$, each {UE}-$i$ will generate a content request ${F^i}(t)$ at an arrival rate $\lambda_{i}\left(t\right)$ which is considered time-varying to be more closely aligned with reality and $0\leq\lambda_{i}\left(t\right)\leq1$. Let ${F^i}(t) \in \emptyset$ denote that {UE}-$i$ generate no content request at time $t$;otherwise $F^{i}(t)\in\mathcal{F}$ when $F^{i}(t)\notin\emptyset$. When $F^{i}(t)\in\mathcal{F}$, the probability of each content $F_{n}\in\mathcal{F}$ requested by {UE}-$i$ at time $t$ is assumed to follow a Zipf distribution \cite{Sade17}, defined as ${\bf P}^{i}\left(\alpha^{i}\left(t\right),t\right)=\left\{ P_{n}^{i}\left(\alpha^{i}\left(t\right),t\right)\right\} _{n=1}^{N}$. The distribution parameter $\alpha^{i}\left(t\right)$ evolves dynamically over time in this paper and is relevant to the subject interests of {UE}-$i$. If ${F^i}(t)$ is uncached in {UE}-$i$, which is represented as $F^{i}\left(t\right)\notin\mathcal{C}_{i}(t)$ and $\mathcal{C}_{i}(t)$ is the contents set cached in {UE}-$i$ at time $t$, {UE}-$i$ will upload this request information to access the absent content from the MEC server. Subsequently, MEC server will search for the requested contents from UEs in its current cache state $\mathcal{C}_{0}(t)$. When $F^{i}\left(t\right)\notin\mathcal{C}_{0}(t)$ happens, the MEC server will further access the absent contents from the cloud. Finally, the absent contents of UEs will be sent back from the MEC server. Note that, the content request $F^{i}\left(t\right)$ can be directly satisfied by the local {UE}-$i$ when $F^{i}\left(t\right)\in\mathcal{C}_{i}(t)$, and the request information will not be uploaded to the MEC server at that time.

\subsection{Local and Global Popularity}\label{lgpopu}

\begin{figure}[!htpb]
\centering
\includegraphics[width=0.48\textwidth]{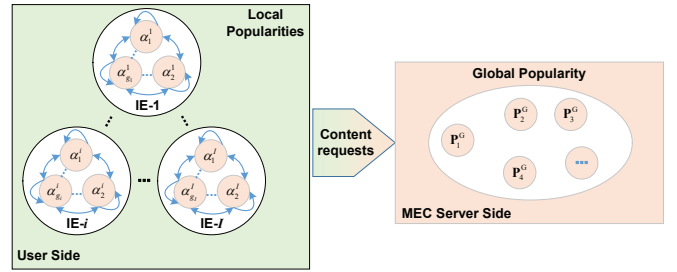}
\caption{Local and global popularity.}
\label{fig2}
\end{figure}

We introduce the local popularity and global popularity to model the time-varying content popularities  depicted in Fig. \ref{fig2}. The local popularity of each {UE} depends on the subjective interests of itself, and the global popularity reflects the comprehensive interest across the service region of the MEC server. With regard to the local popularity, we model the dynamics of ${\alpha ^i}(t)$ using a model-free Markov chain with $\left| {{{\cal G}_i}} \right|$ states recorded in the set ${{\cal G}_i} = \{ \alpha _{g_i}^i|g_i = 1,2, \cdots ,{G_i}\}$, where the ${{{\cal G}_i}}$ as well as the corresponding transition probabilities of ${{{\cal G}_i}}$ are completely unavailable due to the complexity and diversity of subjective interests \cite{Zheng20b}. Moreover, {instead of conventional independent and identically distributed (IID) assumption, we assume less restrictive condition, i.e.,} the behaviors of UEs are independent but not identically distributed. Specifically in our model, the state set ${{\cal G}_i}$ of each {UE}-$i$ as well as the potential state transition probabilities are different and independent. The global popularity at the MEC server side at time $t$ can be denoted as ${{\bf{P}}^{\rm{G}}}(t) = [P_n^{\rm{G}}(t)]_{n = 1}^N$, where $P_n^{\rm{G}}(t)$ is the probability that content $n$ is requested within the entire service area at time $t$.

{\begin{remark}\label{remark1}
Note that if data are processed in an insufficiently random manner, independence can be easily violated due to spatiotemporal correlations. On the other hand, non-identical user behaviors alone can be categorized into many different types, including feature/label distribution skew, concept drift, quantity skew, etc. Additionally, UE and data distributions can fluctuate over time, which compounds the non-IIDness. Learning from highly skewed non-IID data requires characterizing and/or mitigating each of the above effects and even a mixture of them. Although several solutions have been proposed such as data-sharing and model traveling, dealing with real-world non-IID user behaviors still remains a open problem \cite{Kair20}.
\end{remark}}

\subsection{Dynamic Caching Mechanism}

Assume that the MEC server received the request information ${\bf{R}}^{\rm{G}}(t) = \left\{{{ F}^i}(t)\right\}_{i = 1}^I$ from UEs at time $t$, which is the stack of all the absent files at the user side. Then, the MEC server will check its current cache $\mathcal{C}_{0}(t)$ and access to the cloud to get the absent files $\mathcal{C}_{0}^{\textrm{u}}(t)=\left\{ F^{i}\left(t\right)\mid i=1,\cdots I\right\} -\mathcal{C}_{0}(t)$. $\mathcal{C}_{0}^{\textrm{u}}(t)$ will be forwarded to the UEs from the cloud via the server. Therefore, $\mathcal{C}_{0}^{\textrm{u}}(t)$ are the new input files for the MEC server at every time $t$. Additionally, $\mathcal{C}_{0}^{\textrm{u}}(t)$ could be an empty set when the cache hit rate of MEC server at time $t$ reaches $100\%$ . It is worthy to note that ${\bf{R}}^{\rm{G}}(t)$ will be erased from the server before the next time slot by the privacy-preserving mechanism. In addition, to improve the utilization of caching resource, we adopt the dynamic caching policy presented in \cite{Zheng20}. Let $\mathbf{a}_{0}^{-}(t)=\left[a_{c_{0}}^{-}\left(t\right)\right]_{c_{0}=1}^{M_{0}}$ decide which files in $\mathcal{C}_{0}(t)$ should be evicted from MEC server at time $t$, where $a_{c_{0}}^{-}\left(t\right)=1$ indicates that file $F_{c_{0}}^{0}\left(t\right)\in\mathcal{C}_{0}(t)$ should be deleted; otherwise if $a_{c_{0}}^{-}\left(t\right)=0$, it should continue to be retained. Moreover, let $\mathbf{a}_{0}^{+}(t)=\left[a_{c_{0}^{\textrm{u}}}^{+}\left(t\right)\right]_{c_{0}^{\textrm{u}}=1}^{\left|\mathcal{C}_{0}^{\textrm{u}}(t)\right|}$ denote which files in $\mathcal{C}_{0}^{\textrm{u}}(t)$ should be preserved in MEC cache at time $t$, where $a_{c_{0}^{\textrm{u}}}^{+}\left(t\right)=1$ means that file $F_{c_{0}^{\textrm{u}}}^{0}\left(t\right)\in\mathcal{C}_{0}^{\textrm{u}}(t)$ should be stored; otherwise if $a_{c_{0}^{\textrm{u}}}^{+}\left(t\right)=0$, it should be outright discarded. To maximize the utilization of cache resource, we assume that $\left|\mathcal{C}_{0}(t)\right|=M_{0}$. As such, limited by the cache capacity of the MEC server, we have
\begin{equation} \label{e2}
\sum\nolimits _{c_{0}^{\textrm{u}}=1}^{\left|\mathcal{C}_{0}^{\textrm{u}}(t)\right|}a_{c_{0}^{\textrm{u}}}^{+}\left(t\right)=\sum\nolimits _{c_{0}=1}^{M_{0}}a_{c_{0}}^{-}\left(t\right).
\end{equation}
It should be emphasized that the dimension of $\mathbf{a}_{0}^{+}(t)$ is equal to $\left|\mathcal{C}_{0}^{\textrm{u}}(t)\right|$ which is a variable with respect to time $t$. Under this dynamic caching mechanism, the cache state of the MEC server is time-varying and the update operation only happens when new files arrive.

Similarly, this dynamic caching mechanism will be executed in each {UE}. We define the cache deletion of {UE}-$i$ as $\mathbf{a}_{i}^{-}(t)=\left[\mathbf{\mathit{a}}_{c_{i}}^{-}\left(t\right)\right]_{c_{i}=1}^{M_{i}}$, where $\mathbf{\mathit{a}}_{c_{i}}^{-}\left(t\right)=1$ indicates that file $F_{c_{i}}^{i}\left(t\right)\in\mathcal{C}_{i}\left(t\right)$ should de discarded from {UE}-$i$ at time $t$; otherwise if $\mathbf{\mathit{a}}_{c_{i}}^{-}\left(t\right)=0$ , it should be retained in memory. Furthermore, we denote the new file entered into {UE}-$i$ at time $t$ with $\mathcal{C}_{i}^{\textrm{u}}(t)=\left\{ F^{i}\left(t\right)\right\} -\mathcal{C}_{i}(t)$. Obviously, $0\leq\left|\mathcal{C}_{i}^{\textrm{u}}(t)\right|\leq1$. We can also obtain the following cache capability constraint of {UE}-$i$
\begin{equation} \label{e3}
\sum\nolimits _{c_{i}^{\textrm{u}}=1}^{\left|\mathcal{C}_{i}^{\textrm{u}}(t)\right|}a_{c_{i}^{\textrm{u}}}^{+}\left(t\right)=\sum\nolimits _{c_{i}=1}^{M_{i}}a_{c_{i}}^{-}\left(t\right),
\end{equation}
where $a_{c_{i}^{\textrm{u}}}^{+}\left(t\right)$ decides whether file $F_{c_{i}^{\textrm{u}}}^{i}\left(t\right)\in\mathcal{C}_{i}^{\textrm{u}}(t)$ should be preserved in {UE}-$i$ at time $t$ or not. $F_{c_{i}^{\textrm{u}}}^{i}\left(t\right)$ should be stored when $a_{c_{i}^{\textrm{u}}}^{+}\left(t\right)=1$; otherwise $a_{c_{i}^{\textrm{u}}}^{+}\left(t\right)=0$ means file $F_{c_{i}^{\textrm{u}}}^{i}\left(t\right)$ should be discarded or $\left|\mathcal{C}_{i}^{\textrm{u}}(t)\right|=0$ happens. It is worth to mention that the cache preservation indicator of {UE}-$i$ is a scalar resulting from $0\leq\left|\mathcal{C}_{i}^{\textrm{u}}(t)\right|\leq1$, denoted as $a_{i}^{+}\left(t\right)=a_{c_{i}^{\textrm{u}}}^{+}\left(t\right)$.

\subsection{Realtime Cache Hit Rate}

At each time $t$, the MEC server will received a certain amount of requests from the UEs within the service coverage, denoted as $N_{0}^{\textrm{R}}\left(t\right)=\left|\mathbf{R}^{\textrm{G}}\left(t\right)\right|$. Considering the existence of $\lambda_{i}\left(t\right)$ is a variable with $t$ and $0\leq\lambda_{i}\left(t\right)\leq1$, we have $N_{0}^{\textrm{R}}\left(t\right)\leq I$. Then we define the global realtime cache hit rate at the MEC server side as
\begin{equation} \label{e4}
H_{0}\left(t\right)=1-\frac{\left|\mathcal{C}_{0}^{\textrm{u}}(t)\right|}{N_{0}^{\textrm{R}}\left(t\right)}.
\end{equation}
For {UE}-$i$, we define the realtime cache hit rate as
\begin{equation} \label{e5}
H_{i}\left(t\right)=1-\left|\mathcal{C}_{i}^{\textrm{u}}(t)\right|.
\end{equation}
Considering that one {UE} only requests at most one content in a single time slot, $H_{i}\left(t\right)$ can only be equal to 0 or $1$. Here, the sliding average of $H_{i}\left(t\right)$ over a period of time $T_{\textrm{h}}$ is given by
\begin{equation} \label{e6}
H_{i}^{\textrm{savg}}\left(t\right)=\frac{1}{T_{\textrm{h}}}\sum\nolimits _{t_{\textrm{h}}=0}^{T_{\textrm{h}}-1}H_{i}\left(t-t_{\textrm{h}}\right).
\end{equation}

\section{PROBLEM FORMULATION AND ANALYSIS}
\label{sec3}

\subsection{Problem Formulation}

To effectively leverage the limited caching resources in the MEC system, we maximize the distributed cache hit rate of all the devices by optimizing the dynamic caching mechanism within the constraint of privacy preservation. Furthermore, we maximize the long-term cache hit rate over a continuous period of time. Therefore, the underlying optimization problem at the MEC side is formulated as follows:
\begin{subequations}\label{e7}
    \begin{align}
    P_1:\quad
    \underset{\mathcal{A}_{0}}{\max} \quad
    &\underset{\Gamma\rightarrow\infty}{\textrm{lim}}\sum\nolimits_{\tau=0}^{\Gamma}\mathbb{E\mathit{\left[\chi^{\tau}H_{0}\left(t+\tau\right)\right]}},\label{e7a}\\
    {\rm{s.t.}}\quad \quad
    & \eqref{e2}, \label{e7b}\\
    &\left|\mathcal{C}_{0}(t)\right|\leq M_{0},\label{e7c}\\
    &a_{c_{0}}^{-}\left(t\right)\!\in\!\left\{ 0,1\right\}\!, \forall c_{0}\!\in\!\mathcal{M}_{0}, \label{e7d}\\
    &a_{c_{0}^{\textrm{u}}}^{+}\left(t\right)\!\in\!\left\{ 0,1\right\}\!,\forall c_{0}^{\textrm{u}}\!\in\!\mathcal{M}_{0}^{\textrm{u}}\left(t\right),\label{e7e}
    \end{align}
\end{subequations}
where $\mathcal{A}_{0}\!=\!\left\{\mathbf{a}_{0}\left(t\right)\!=\!\left(\mathbf{a}_{0}^{+}\left(t\right)\!,\mathbf{a}_{0}^{-}\left(t\right)\right)\!\mid \!t\!=\!0,1,2,\!\cdots\!\right\}$ represents the collection of dynamic caching actions at the MEC server side in each time slot $t$. $\chi\in\left[0,1\right]$ is the discount factor, and the expectation is taken with respect to the measure included by the decision variables as well as the system state. Besides, $\mathcal{M}_{0}^{\textrm{u}}\left(t\right)=\left\{ 1,2,\cdots\left|\mathcal{C}_{0}^{\textrm{u}}(t)\right|\right\}$ and $\mathcal{M}_{0}=\left\{ 1,2,\cdots,M_{0}\right\}$. The constraint in \eqref{e7c} reflects the limitation of caching capability of the MEC server, and \eqref{e7b} ensures a balance in the size of the cached files at the MEC server after the caching replacement to keep the cache full but not overflowed. At the local user side, the optimization problem of arbitrary {UE}-$i$ can be formulated as
\begin{subequations}\label{e8}
    \begin{align}
    P_2:\quad
    \underset{\mathcal{A}_{i}}{\textrm{max}} \quad
    &\underset{\Gamma\rightarrow\infty}{\textrm{lim}}\sum\nolimits_{\tau=0}^{\Gamma}\mathbb{E\mathit{\left[\chi^{\tau}H_{i}^{\textrm{savg}}\left(t+\tau\right)\right]}},\label{e8a}\\
    {\rm{s.t.}}\quad \quad
    & \eqref{e3}, \label{e8b}\\
    &\left|\mathcal{C}_{i}(t)\right|\leq M_{i},\label{e8c}\\
    &a_{c_{i}}^{-}\left(t\right)\!\in\!\left\{ 0,1\right\}\!,\forall c_{i}\!\in\!\mathcal{M}_{i}, \label{e8d}\\
    &a_{c_{i}^{\textrm{u}}}^{+}\left(t\right)\!\in\!\left\{ 0,1\right\}\!,\forall c_{i}^{\textrm{u}}\!\in\!\mathcal{M}_{i}^{\textrm{u}}\left(t\right),\label{e8e}
    \end{align}
\end{subequations}
where $\mathcal{A}_{i}\!=\!\left\{ \mathbf{a}_{i}\left(t\right)\!=\!\left(a_{i}^{+}\left(t\right)\!,\mathbf{a}_{i}^{-}\left(t\right)\right)\!\mid \! t\!=\!0,1,2,\!\cdots\!\right\}$ is the collection of dynamic caching actions on {UE}-$i$ in each time slot $t$. Besides, $\mathcal{M}_{i}^{\textrm{u}}\left(t\right)=\left\{ 1,2,\cdots\left|\mathcal{C}_{i}^{\textrm{u}}(t)\right|\right\}$ and $\mathcal{M}_{i}=\left\{ 1,2,\cdots,M_{i}\right\}$.

The following facts and technical challenges of problems \eqref{e7} and \eqref{e8} should be noted:
\begin{itemize}
  \item The objective functions of the problems are both accumulated over time rather than instantaneous functions.
  \item The solutions of problem \eqref{e7} and \eqref{e8} are both dynamic strategy over time rather than a transient one. Moreover, the dimension of $\mathbf{a}_{0}^{+}\left(t\right)$ is time-varying.
  \item The cache states and actions of the MEC server and the UEs conform to contextual chain property over time.
  \item The distributed problems formulated above are interactional but the privacy-preserving mechanism prevents the information exchange among the problems.
\end{itemize}

\subsection{Problem Recast}

To overcome the first two technical challenges as well as considering the fact of the chain property mentioned in the third, we convert the underlying optimization problem into a Markov decision process (MDP) which consists of four components, i.e, state space, action space, state transition probabilities, and reward. Specifically, the MDP descriptions of the problems \eqref{e7} and \eqref{e8} are denoted as $\left\langle \mathcal{S}_{0},\mathcal{A}_{0},\mathbf{P}_{0},R_{0}\right\rangle$ and $\left\langle \mathcal{S}_{i},\mathcal{A}_{i},\mathbf{P}_{i},R_{i}\right\rangle$ respectively.

\subsubsection{States}

Considering the time variable $t$, $\mathcal{S}_{0}$ and $\mathcal{S}_{i}$ actually can be denoted as $\mathcal{S}_{0} = \left\{\mathbf{s}_{0}\left(t\right) \mid t = 0,1,2,\cdots\right\}$ and $\mathcal{S}_{i} = \left\{\mathbf{s}_{i}\left(t\right) \mid t = 0,1,2,\cdots\right\}$ respectively. According to the necessary information required by the dynamic caching actions, we define $\mathbf{s}_{0}\left(t\right)=\left\{\mathcal{C}_{0}(t),\mathbf{R}^{\textrm{G}}\left(t\right)\right\}$ and $\mathbf{s}_{i}\left(t\right)=\left\{ \mathcal{C}_{i}(t),\mathbf{R}^{i}\left(t\right)\right\}$, where ${{\bf{R}}^i}(t) = [{{ F}^i}(t - H), \cdots ,{{ F}^i}(t)]$ is a extractor of {UE}-$i$ to extract its historical requests of continuous $H$ times before time $t$. $H$ is the observation window length of the extractor.

\subsubsection{Actions}

From the distributed problems formulated above, we already have $\mathcal{A}_{0}$ and $\mathcal{A}_{i}$, $\forall i\in\mathcal{I}$.

\subsubsection{State Transition}

State transition probability describes that the system transits from one state to the next state under current actions. For problems \eqref{e7} and \eqref{e8}, the state transition probability can be respectively denoted as $P_{\mathbf{s}_{0}\left(t\right)\rightarrow\mathbf{s}_{0}\left(t+1\right)}^{\mathbf{a}_{0}\left(t\right)}$ and $P_{\mathbf{s}_{i}\left(t\right)\rightarrow\mathbf{s}_{i}\left(t+1\right)}^{\mathbf{a}_{i}\left(t\right)}$. However in our problems, $\mathbf{R}^{i}\left(t\right)$ and $\mathbf{R}^{0}\left(t\right)$ depend on the local and global popularity described in Section \ref{lgpopu}, which results in the transition probability unavailable.

\subsubsection{Reward}

The reward function assigns each perceived state to a value associated with an explicit goal. For an MDP, when an action is taken under a state, the state will transfer to next state and the environment will return an instantaneous reward as a feedback immediately, which is respectively derived as the cache hit rate $H_{0}\left(t\right)$ and $H_{i}^{\textrm{savg}}\left(t\right)$ in our problems. On this basis, the cumulative reward starting from time $t$ can be respectively given by
\begin{equation} \label{e8.1}
r_{0}\left(t\right)=\sum\nolimits_{\tau=0}^{\Gamma}\chi^{\tau}H_{0}\left(t+\tau\right),
\end{equation}
\begin{equation} \label{e8.2}
r_{i}\left(t\right)=\sum\nolimits_{\tau=0}^{\Gamma}\chi^{\tau}H_{i}^{\textrm{savg}}\left(t+\tau\right),
\end{equation}

Specifically in our problems, the critical component $\mathcal{S}_{0}$ of a MDP is unobservable under the privacy-preserving mechanism. The reason is that, $\mathbf{R}^{\textrm{G}}\left(t\right)$ as a component of $\mathbf{s}_{0}\left(t\right)$ is the privacy of the UEs and must be immediately erased from the MEC server in current time slot. Thus, the MEC server cannot observe $\mathbf{s}_{0}\left(t\right)$ at any time $t$, which leads to $\mathcal{S}_{0}$ unavailable. Therefore, the technical bottlenecks from the fourth challenge still remain, especially for the MDP problem converted from problem \eqref{e7}.

\subsection{Privacy-Preserving Distributed Popularity Prediction}

To allow privacy preservation as well as help all devices cache contents more effectively, we herein introduce the local and global popularity into the system states. In detail, we replace $\mathbf{R}^{i}\left(t\right)$ in $\mathbf{s}_{i}\left(t\right)$ and $\mathbf{R}^{\textrm{G}}\left(t\right)$ in $\mathbf{s}_{0}\left(t\right)$ with the future contents popularity ${{\bf{P}}^i}({\alpha ^i}(t+1),t+1)$ and ${{\bf{P}}^{\rm{G}}}(t+1)$ respectively, renewed as
\begin{equation} \label{e9}
\mathbf{\widetilde{s}}_{i}\left(t\right)=\left\{ \mathcal{C}_{i}(t),{\bf P}^{i}\left(\alpha^{i}\left(t+1\right),t+1\right)\right\},
\end{equation}
\begin{equation} \label{e10}
\mathbf{\widetilde{s}}_{0}\left(t\right)=\left\{ \mathcal{C}_{0}(t),{\bf P}^{{\rm G}}\left(t+1\right)\right\}.
\end{equation}
The state space can be accordingly rewritten as $\mathcal{\widetilde{S}}_{0}=\left\{\mathbf{\widetilde{s}}_{0}\left(t\right)\mid t=0,1,2,\cdots\right\}$ and $\mathcal{\widetilde{S}}_{i}=\left\{\mathbf{\widetilde{s}}_{i}\left(t\right)\mid t=0,1,2,\cdots\right\}$

As clarified earlier, the variation of ${\bf P}^{i}\left(\alpha^{i}\left(t+1\right),t+1\right)$ and ${\bf P}^{{\rm G}}\left(t+1\right)$ depend on the interests of UEs which is subjective and complicated. Thus, ${\bf P}^{i}\left(\alpha^{i}\left(t+1\right),t+1\right)$ and ${\bf P}^{{\rm G}}\left(t+1\right)$ are unobservable especially under the constraint of privacy preservation. Here, we introduce a FL method to predict the dynamic popularities while preserving user privacy. Specifically, we deploy the prediction model with the same architecture of neural network on each device in the system. At the local user side, the future popularity's prediction of {UE}-$i$ is based on the historical requests reserved in its equipment and the prediction can be denoted as
\begin{equation} \label{e11}
{{{\bf{\hat P}}}^i}({\alpha ^i}(t + 1),t + 1) = {f^{{\Theta ^i}}}({{\bf{R}}^i}(t)),
\end{equation}
where ${f^{{\Theta ^i}}}( \cdot )$ is the local predictive model in {UE}-$i$ and ${\Theta ^i}$ is the collection of trainable parameters. ${{{\bf{\hat P}}}^i}({\alpha ^i}(t + 1),t + 1)$ is the prediction of ${{\bf{P}}^i}({\alpha ^i}(t + 1),t + 1)$. At the MEC server side at time $t$, the temporary ${\bf{R}}^{\rm{G}}(t)$ can be used by the URFL method for global prediction before the erase operation, which can be denoted as
\begin{equation} \label{e15}
{{{\bf{\hat P}}}^{\rm{G}}}(t + 1) = f^{{\Theta ^{\rm{G}}}}({\bf{R}}^{\rm{G}}(t)),
\end{equation}
where ${{{\bf{\hat P}}}^{\rm{G}}}(t)$ denotes the prediction of global popularity ${{\bf{P}}^{\rm{G}}}(t+1)$. $f^{{\Theta ^{\rm{G}}}}( \cdot )$ is the global predictive model in the MEC server. ${{\Theta ^{\rm{G}}}}$ is the parameters set.

To train these prediction models under privacy preservation, the FL framework is adopted. At the local user side, the database formed by ${{\bf{R}}^i}({t})$ is used for the local training of ${\Theta ^i}$ and the connectivity between UEs is not existing. At the MEC server side, the parameters set ${{\Theta ^{\rm{G}}}}$ is obtained by the parameters aggregation based on the FL framework, which can be denoted as
\begin{equation}
{\Theta ^{\rm{G}}}{\rm{ = }}\frac{1}{I}\sum\nolimits_{i = 1}^I {{\omega _i}\Theta^i}.
\end{equation}
where ${{\omega _i}}$ is the aggregation weight and ${\Theta ^i}$ is uploaded by the {UE}-$i$ every a certain local training step. Once a weight aggregation is complete, the new parameters ${\Theta ^{\rm{G}}}$ will be broadcast to all UEs for a new round of local training until the models converged. Because the local training is performed alone on its local equipment and the interaction between the local UEs and the MEC server only involves the passing of prediction model parameters, the user privacy, i.e., ${{\bf{R}}^i}({t})$, is thus preserved during this training phase.

\begin{figure*}[!htbp]
\newcounter{MYtempeqncnt}
\setcounter{MYtempeqncnt}{\value{equation}}
\setcounter{equation}{14}
\begin{equation} \label{e16.1}
\begin{array}{l}
\pi_{0}^{*} = \mathop{\arg\max}\limits_{\mathbf{a}_{0}\left(t\right)\in\mathcal{A}_{0}}
\displaystyle\sum\nolimits _{\mathbf{\widetilde{s}}_{0}\left(t+1\right)\in\mathcal{\widetilde{S}}_{0}}P_{\mathbf{\widetilde{s}}_{0}\left(t\right)\to\mathbf{\widetilde{s}}_{0}\left(t+1\right)}^{\mathbf{a}_{0}\left(t\right)}\left(H_{0}(t)+\chi V^{\pi_{0}}\left(\mathbf{\widetilde{s}}_{0}\left(t+1\right)\right)\right),
\end{array}
\end{equation}
\begin{equation} \label{e16.2}
\begin{array}{l}
\pi_{i}^{*} = \mathop{\arg\max}\limits _{\mathbf{a}_{i}\left(t\right)\in\mathcal{A}_{i}}
\displaystyle\sum\nolimits _{\mathbf{\widetilde{s}}_{i}\left(t+1\right)\in\mathcal{\widetilde{S}}_{i}}P_{\mathbf{\widetilde{s}}_{i}\left(t\right)\to\mathbf{\widetilde{s}}_{i}\left(t+1\right)}^{\mathbf{a}_{i}\left(t\right)}\left(H_{i}^{\textrm{savg}}\left(t\right)+\chi V^{\pi_{i}}\left(\mathbf{\widetilde{s}}_{i}\left(t+1\right)\right)\right),
\end{array}
\end{equation}
\setcounter{equation}{\value{MYtempeqncnt}}
\hrulefill
\end{figure*}

\begin{figure*}[!htbp]
\centering
\includegraphics[width=0.92\textwidth]{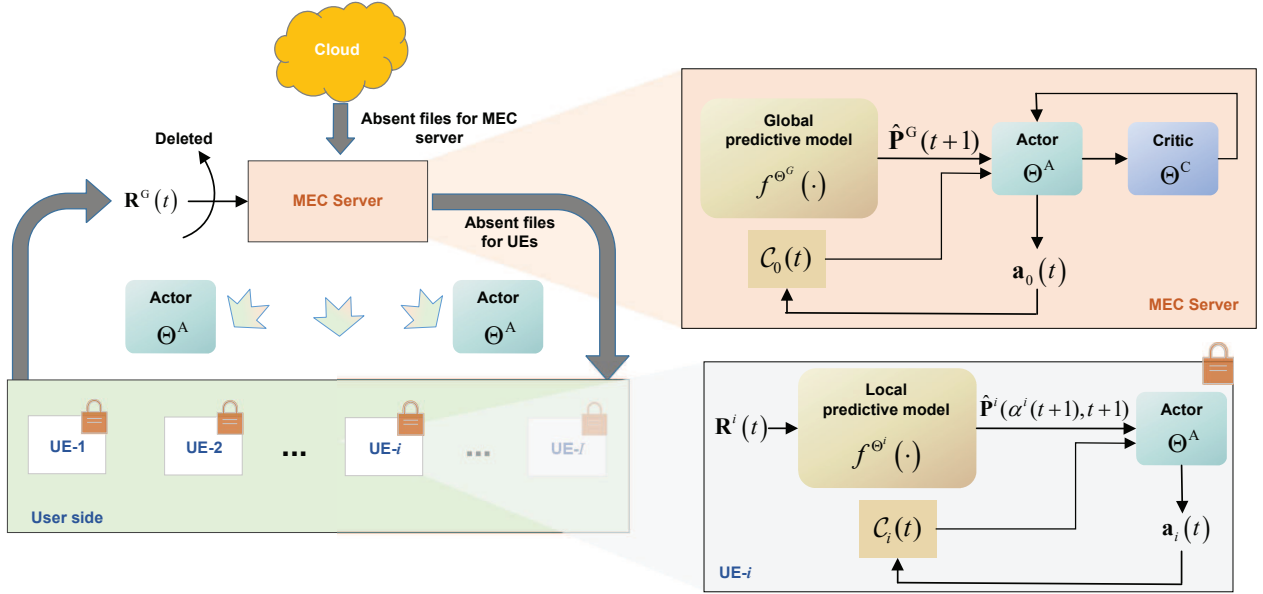}
\caption{Framework of P2D3PG algorithm.}
\label{P2D3PG}
\end{figure*}

After the distributed popularity prediction, the challenges posed by the unobservable state space $\mathcal{S}_{0}$ has been addressed. Then, the optimal policy $\pi_{0}^{*}$ and $\pi_{i}^{*}$ for problem \eqref{e7} and \eqref{e8} can be  respectively derived as equations \eqref{e16.1} and \eqref{e16.2} based on the Bellman's equation, where $V^{\pi_{0}}\left(\mathbf{\widetilde{s}}_{0}\left(t+1\right)\right)=r_{0}\left(t+1\right)$ is the value function under policy $\pi_{0}$ at sate $\mathbf{\widetilde{s}}_{0}\left(t+1\right)$, and $V^{\pi_{i}}\left(\mathbf{\widetilde{s}}_{i}\left(t+1\right)\right)=r_{i}\left(t+1\right)$ is the value function under policy $\pi_{i}$ at sate $\mathbf{\widetilde{s}}_{i}\left(t+1\right)$. Whereas, according to the local and global popularity model in our system, it can be found that the $P_{\mathbf{\widetilde{s}}_{0}\left(t\right)\rightarrow\mathbf{\widetilde{s}}_{0}\left(t+1\right)}^{\mathbf{a}_{0}\left(t\right)}$ and $P_{\mathbf{\widetilde{s}}_{i}\left(t\right)\rightarrow\mathbf{\widetilde{s}}_{i}\left(t+1\right)}^{\mathbf{a}_{i}\left(t\right)}$ still can not be acquired even if we get the ${{{\bf{\hat P}}}^i}({\alpha ^i}(t + 1),t + 1)$ and ${{{\bf{\hat P}}}^{\rm{G}}}(t + 1)$. As such, traditional optimization techniques such as dynamic programming cannot effectively solve our problems, and we will propose a privacy-preserving distributed reinforcement learning algorithm to solve this problems.

\section{P2D3PG FOR DYNAMIC EDGE CACHING}\label{method}

Once ${{{\bf{\hat P}}}^{\rm{G}}}(t + 1)$ is predicted, certain EC policy should be subsequently determined and implemented to maximize the EC hit rate of the entire MEC system. In this work, we propose a P2D3PG algorithm for this purpose, and the designed algorithm framework is illustrated in Fig. \ref{P2D3PG}.

\subsection{MEC Server Side}

First, the MEC server receives the requests information $\mathbf{R}^{\textrm{G}}\left(t\right)$ from UEs at the beginning of each time slot $t$. Subsequently, $\mathbf{R}^{\textrm{G}}\left(t\right)$ is fed into the global predictive model obtained by URFL to predict the global popularity ${{{\bf{\hat P}}}^{\rm{G}}}(t + 1)$ of next time slot $t+1$ based on equation \eqref{e15}. Meanwhile, the absent files of UEs will be delivered to UEs while the $\mathbf{R}^{\textrm{G}}\left(t\right)$ is immediately erased from the MEC server in time slot $t$. Then combining ${{{\bf{\hat P}}}^{\rm{G}}}(t + 1)$ with the current cache state of the MEC server $\mathcal{C}_{0}(t)$, the state $\mathbf{\widetilde{s}}_{0}\left(t\right)$ can be obtained. Subsequently $\mathbf{\widetilde{s}}_{0}\left(t\right)$ is fed into the actor network, which is also a neural network with several dense layers.
The actor network equals to a parameterized actor function $\mathbf{a}_{0}\left(t\right)=\pi^{\Theta^{\textrm{A}}}\left(\mathbf{\widetilde{s}}_{0}\left(t\right)\right)$ which specifies the current policy by deterministically mapping states to a specific action, where $\pi$ represents a policy on parameters $\Theta^{\textrm{A}}$. In order for the agent to fully explore the environment, exploration-exploitation method is adopted. Different from the $\varepsilon$-greedy exploration \cite{Sutt98} which is effective for small or discrete action space. In this work, we balance the exploration and the exploitation by adding a gaussian noise vector on the policy output, i.e
\setcounter{equation}{16}
\begin{equation} \label{e19}
\mathbf{\widetilde{a}}_{0}\left(t\right)=\pi^{\Theta^{\textrm{A}}}\left(\mathbf{\widetilde{s}}_{0}\left(t\right)\right)+\left.\mathbf{n_{0}}\left(t\right)\right|_{n_{0}\left(t\right)\sim\mathcal{N}\left(0,\sigma^{2}\right)},
\end{equation}
where $\mathbf{n_{0}}\left(t\right)$ is the gaussian noise vector and $n_{0}\left(t\right)$ is the component following a gaussian distribution with a mean of 0 and a variance of $\sigma^{2}$. Then the action $\mathbf{\widetilde{a}}_{0}\left(t\right)$ will be sent to the critic network which is also a neural network containing several dense layers together with the state $\mathbf{\widetilde{s}}_{0}\left(t\right)$. Consequently, the critic network will output the estimate of the target-Q value $Q\left(\left.\mathbf{\widetilde{s}}_{0}\left(t+1\right),\mathbf{\widetilde{a}}_{0}\left(t+1\right)\right|\Theta^{\textrm{C}}\right)$ which is a step forward for estimating the Q-value defined as \eqref{e20}.
\begin{equation} \label{e20}
\begin{array}{l}
Q\left(\left.\mathbf{\widetilde{s}}_{0}\left(t\right),\mathbf{\widetilde{a}}_{0}\left(t\right)\right|\Theta^{\textrm{C}}\right)=\\
\quad \mathbb{E}_{\pi_{0}}\left[\left.\sum\nolimits_{\tau=0}^{+\infty}\chi^{\tau}H_{0}\left(t+\tau\right)\right|\mathbf{\widetilde{s}}_{0}\left(t\right),\mathbf{\widetilde{a}}_{0}\left(t\right)\right].
\end{array}
\end{equation}
$\Theta^{\textrm{C}}$ is the trainable parameters of the critic network. After a further linear transformation, the output $Q\left(\left.\mathbf{\widetilde{s}}_{0}\left(t+1\right),\mathbf{\widetilde{a}}_{0}\left(t+1\right)\right|\Theta^{\textrm{C}}\right)$ will be fed back to the actor network while contributes to the loss function of the actor. In addition, the cache state of the MEC server $\mathcal{C}_{0}(t)$ at time $t$ will be updated to the next cache state $\mathcal{C}_{0}(t+1)$ following the guidance of the action $\mathbf{\widetilde{a}}_{0}\left(t\right)$.

In practical training, the two networks $\pi^{\Theta^{\textrm{A}}}\left(\cdot\right)$ and $Q\left(\left.\cdot\right|\Theta^{\textrm{C}}\right)$ are called online networks. Correspondingly for a stabler and faster convergence, there are two counterparts respectively called target actor network $\pi^{\overline{\Theta}^{\textrm{A}}}\left(\cdot\right)$ and target critic network $Q\left(\cdot\left|\overline{\Theta}^{\textrm{C}}\right.\right)$ whose architectures and parameters are clone from their online networks every a few steps.

\begin{algorithm}[!htbp]
\caption{P2D3PG for dynamic EC at the MEC server.}
\label{alg_P2D3PG_mec}
\begin{algorithmic}[1]
 \STATE  \textbf{Initialize:} Initialize $\Theta^{\textrm{A}}$, $\Theta^{\textrm{C}}$ and memory buff $\Omega$. Obtain the initial $\overline{\Theta}^{\textrm{A}}$ and $\overline{\Theta}^{\textrm{C}}$ by cloning $\Theta^{\textrm{A}}$ and $\Theta^{\textrm{C}}$.
\STATE \textbf{For} episode $=1,2,\cdots,\varPsi$ \textbf{MEC do:}
\STATE \quad Initialize cache state $\mathcal{C}_{0}(0)$. Initialize $\mathbf{R}^{\textrm{G}}\left(0\right)$.
\STATE \quad \textbf{For} $t=1,2,\cdots,\Upsilon$ \textbf{do:}
\STATE \quad \quad Receive $\mathbf{R}^{\textrm{G}}\left(t\right)$ from UEs. Then predict $\hat{{\bf P}}^{{\rm G}}\left(t+1\right)$\\
       \quad \quad by \eqref{e15} under the proposed URFL.
\STATE \quad \quad Observe the state $\mathbf{\widetilde{s}}_{0}\left(t\right)$, and observe the reward\\
       \quad \quad feedback $H_{0}\left(t\right)$ by \eqref{e4}.
\STATE \quad \quad Delete $\mathbf{R}^{\textrm{G}}\left(t\right)$ for privacy preservation.
\STATE \quad \quad Update $\mathcal{C}_{0}(t)$ to $\mathcal{C}_{0}(t+1)$ under action $\mathbf{\widetilde{a}}_{0}\left(t\right)$ by \eqref{e19}.
\STATE \quad \quad Store point $\left(\mathbf{\widetilde{s}}_{0}\left(t-1\right),\mathbf{\widetilde{a}}_{0}\left(t\right),H_{0}\left(t\right),\mathbf{\widetilde{s}}_{0}\left(t\right)\right)$ in $\Omega$.
\STATE \quad \quad Randomly sample a mini-batch of $N_{\textrm{s}}$ points from $\Omega$.
\STATE \quad \quad Calculate $y\left(t_{n_{\textrm{s}}}\right)$ by \eqref{e22}. Then update $\Theta^{\textrm{C}}$ by \eqref{e21}\\
       \quad \quad and $\nabla_{\Theta^{\textrm{C}}}L\left(\Theta^{\textrm{C}}\right)$. Update $\Theta^{\textrm{A}}$ by \eqref{e25}.
\STATE \quad \quad Soft-update the target actor/critic every $\varphi$ steps:
\[\left\{ \begin{array}{c}
\overline{\Theta}^{\textrm{C}}\leftarrow\nu\Theta^{\textrm{C}}+\left(1-\nu\right)\overline{\Theta}^{\textrm{C}}\\
\overline{\Theta}^{\textrm{A}}\leftarrow\nu\Theta^{\textrm{A}}+\left(1-\nu\right)\overline{\Theta}^{\textrm{A}}
\end{array}\right.\]
\STATE \quad \textbf{End For}
\STATE \textbf{End For}
\end{algorithmic}
\end{algorithm}
During the train phase at the MEC server, we adopt experience replay to enhance the stability of the training. The dataset in the replay buff can be denoted as
\begin{equation}\label{e20.5}
\Omega=\left\{ \left(\mathbf{\widetilde{s}}_{0}\left(t\right)\!,\mathbf{\widetilde{a}}_{0}\left(t\right)\!,H_{0}\left(t\right)\!,\mathbf{\widetilde{s}}_{0}\left(t+1\right)\right)\right\}.
\end{equation}
Specifically during the mini-batch training, $N_{\textrm{s}}$ samples $\left\{ \left(\mathbf{\widetilde{s}}_{0}\left(t_{n_{\textrm{s}}}\right)\!,\mathbf{\widetilde{a}}_{0}\left(t_{n_{\textrm{s}}}\right)\!,H_{0}\left(t_{n_{\textrm{s}}}\right)\!,\mathbf{\widetilde{s}}_{0}\left(t_{n_{\textrm{s}}}+1\right)\right)\right\}$ $(n_{\textrm{s}}\!\in\!\left\{ 1,2,\cdots,N_{\textrm{s}}\right\})$ are randomly taken  as a mini-batch from the replay buffer $\Omega$, where $t_{n_{\textrm{s}}}$ is the random sample point at time $t$. Then we train the actor network and the critic network jointly. To let the critic network $Q\left(\cdot\left|\overline{\Theta}^{\textrm{C}}\right.\right)$ approach the real Q value function which will be further used to guide the training of the actor, the training loss function of the critic in the MSE sense can be defined as
\begin{equation}\label{e21}
L\!\left(\!\Theta^{\textrm{C}}\!\right)\!=\!\frac{1}{N_{\textrm{s}}}\!\sum\limits_{n_{\textrm{s}}=1}^{N_{\textrm{s}}}\left(y\left(t_{n_{\textrm{s}}}\right)\!-\!Q\left(\left.\mathbf{\widetilde{s}}_{0}\left(t_{n_{\textrm{s}}}\right)\!,\mathbf{\widetilde{a}}_{0}\left(t_{n_{\textrm{s}}}\right)\right|\!\Theta^{\textrm{C}}\!\right)\!\right)^{2}\!,
\end{equation}
where
\begin{equation}\label{e22}
y\left(t_{n_{\textrm{s}}}\right)\!=\!H_{0}\left(t_{n_{\textrm{s}}}\right)\!+\!\chi Q\!\left(\!\left.\mathbf{\widetilde{s}}_{0}\left(t_{n_{\textrm{s}}}\!+\!1\right)\!,\mathbf{a}_{0}\left(t_{n_{\textrm{s}}}\!+\!1\right)\right|\overline{\Theta}^{\textrm{C}}\right).
\end{equation}
$t_{n_{\textrm{s}}}$ is the random sample points over time. Thus, we optimize $\Theta^{\textrm{C}}$ by minimizing this MSE loss and $\Theta^{\textrm{C}}$ can be updated by $\nabla_{\Theta^{\textrm{C}}}L\left(\Theta^{\textrm{C}}\right)$. Consequently, $Q\left(\left.\mathbf{\widetilde{s}}_{0}\left(t_{n_{\textrm{s}}}\right),\mathbf{\widetilde{a}}_{0}\left(t_{n_{\textrm{s}}}\right)\right|\Theta^{\textrm{C}}\right)$ will gradually approximate the real Q-value.

\begin{figure*}[!hpbt]
\setcounter{equation}{24}
\begin{equation}\label{e26}
\begin{array}{l}
\nabla_{\Theta^{\textrm{A}}\!}J_{\beta}\left(\pi\right)\approx\displaystyle\frac{1}{N_{\textrm{m}}}\!\displaystyle\sum_{n_{\textrm{m}}=1}^{N_{\textrm{m}}}\!\left(\!\nabla_{\mathbf{a}_{0}}\!\left.Q\!\left(\mathbf{s}_{0}\!\left(t_{n_{\textrm{m}}}\right)\!,\mathbf{\widetilde{a}}_{0}\!\left(t_{n_{\textrm{m}}}\right)\!\left|\Theta^{\textrm{A}}\!\right.\right)\!\right|_{\mathbf{\widetilde{a}}_{0}\left(t_{n_{\textrm{m}}}\right)=\pi^{\Theta^{\textrm{A}}}\!\left(\mathbf{s}_{0}\left(t_{n_{\textrm{m}}}\right)\right)+\mathbf{n}_{0}\left(t_{n_{\textrm{m}}}\right)}\!\cdot\!\nabla_{\!\Theta^{\textrm{A}}}\pi^{\Theta^{\textrm{A}}}\!\left(\mathbf{s}_{0}\!\left(\!t_{n_{\textrm{m}}\!}\right)\right)\!\right).
\end{array}
\end{equation}
\hrulefill
\end{figure*}

The actor is aimed at producing an optimal policy by maximizing the Q-value, denoted as
\setcounter{equation}{21}
\begin{equation}\label{e23}
\pi^{\Theta^{\textrm{A}}}\left(\mathbf{s}_{0}\right)=\textrm{\ensuremath{\mathop{\mathop{\arg\;\max}\limits _{\mathbf{a}_{0}}}}} ~Q\left(\mathbf{s}_{0},\mathbf{a}_{0}\left|\Theta^{\textrm{A}}\right.\right),
\end{equation}
Thus, a performance objective function for the current policy evaluation is designed as
\begin{equation}\label{e24}
J_{\beta}\left(\pi\right)=E_{\mathbf{s}_{0}\sim\rho^{\beta}}\left[Q\left(\mathbf{s}_{0},\mathbf{a}_{0}\left|\Theta^{\textrm{A}}\right.\right)\right],
\end{equation}
which estimates the expectation of $Q\left(\mathbf{s}_{0},\mathbf{a}_{0}\left|\Theta^{\textrm{A}}\right.\right)$ under the state distribution $\mathbf{s}_{0}\sim\rho^{\beta}$. Then, the actor is updated by applying the chain rule to the expected return from the start state distribution with respect to the actor parameters $\Theta^{\textrm{A}}$:
\begin{equation}\label{e25}
\begin{array}{l}
	\nabla_{\Theta^{\textrm{A}}}J_{\beta}\left(\pi\right)=\\
	E_{\mathbf{s}_{0}\sim\rho^{\beta}}\!\left[\nabla_{\mathbf{a}_{0}}\!\left.\!Q\!\left(\mathbf{s}_{0},\mathbf{a}_{0}\!\left|\!\Theta^{\textrm{A}}\!\right.\right)\right|_{\mathbf{a}_{0}\!=\!\pi^{\Theta^{\textrm{A}}}\left(\mathbf{s}_{0}\right)}\!\cdot\!\nabla_{\Theta^{\textrm{A}}}\pi^{\Theta^{\textrm{A}}}\!\left(\mathbf{s}_{0}\right)\right].
\end{array}
\end{equation}
During the practical training, $y\left(t_{n_{\textrm{s}}}\right)$ is sent to the actor network as current real Q-value according to \eqref{e22}. Besides, a mini-batch Monte Carlo sampling with a size of $N_{\textrm{m}}$ is adopted to estimate the expectation, which yields an unbiased estimation shown in \eqref{e26}, where $t_{n_{\textrm{m}}}$ denotes the random sample point at time instant $t$.

\begin{algorithm}[!htbp]
\caption{P2D3PG for dynamic EC at the local UEs.}
\label{alg_P2D3PG_user}
\begin{algorithmic}[1]
\STATE Each {UE}-$i\in\mathcal{I}$ \textbf{in parallel do:}
\STATE \quad Initialize cache state $\mathcal{C}_{i}(0)$ and extractor ${{\bf{R}}^i}(0)$.
\STATE \quad Receive the actor $\Theta^{\textrm{A}}$ broadcasted from the MEC server.
\STATE \quad \textbf{For} $t=0,1,\cdots,\Upsilon$ \textbf{do:}
\STATE \quad \quad Get the historical requests by ${{\bf{R}}^i}(t)$.
\STATE \quad \quad Predict ${\bf \hat{P}}^{i}\left(\alpha^{i}\left(t+1\right),t+1\right)$ by \eqref{e11}
\STATE \quad \quad Observe the state $\mathbf{\widetilde{s}}_{i}\left(t\right)$, and observe the reward\\
       \quad \quad feedback $H_{i}^{\textrm{savg}}\left(t\right)$ by \eqref{e6}.
\STATE \quad \quad Select action $\mathbf{a}_{i}\left(t\right)$ and update $\mathcal{C}_{i}(t)$ to $\mathcal{C}_{i}(t+1)$
\STATE \quad \quad Made the new request ${{ F}^i}(t+1){|_{{{\bf{P}}^i}({\alpha ^i}(t),t)}}$.
\STATE \quad \textbf{End For}
\end{algorithmic}
\end{algorithm}

\subsection{Local User Side}

Furthermore, as illustrated in Fig. \ref{P2D3PG}, the MEC server then broadcasts the trained actor to the UEs within its service coverage. For each {UE}-$i$, the prediction of the future content popularity ${{{\bf{\hat P}}}^i}({\alpha ^i}(t + 1),t + 1)$ should be firstly obtained by feeding ${{\bf{R}}^i}(t)$ into the local predictive model shown in Fig. \ref{P2D3PG}. Then the state $\mathbf{\widetilde{s}}_{i}\left(t\right)$ consisting of $\mathcal{C}_{i}(t)$ and ${{{\bf{\hat P}}}^i}({\alpha ^i}(t + 1),t + 1)$ is fed to the actor which outputs the action $\mathbf{a}_{i}\left(t\right)$, denoted as
\setcounter{equation}{25}
\begin{equation}\label{e27}
\mathbf{a}_{i}\left(t\right)=\pi^{\Theta^{\textrm{A}}}\left(\mathbf{\widetilde{s}}_{i}\left(t\right)\right).
\end{equation}
Following $\mathbf{a}_{i}\left(t\right)$, {UE}-$i$ updates its cache state to $\mathcal{C}_{i}(t+1)$ based on the uncached files $\mathcal{C}_{i}^{\textrm{u}}(t)$ which are accessed from the MEC server. Finally, the request of {UE}-$i$ at time $t$ is satisfied and a new request will be generated subsequently. The overall process of the proposed P2D3PG algorithm at the MEC server and the local UEs is summarized in \textbf{Algorithm \ref{alg_P2D3PG_mec}} and \textbf{Algorithm \ref{alg_P2D3PG_user}}, respectively, where $\varPsi$ is the total episodes, $\varphi$ is the step interval between the online/target networks in parameter clone, and $\nu$ is the coefficient of the soft-update, which is normally set to $0.001$.

Based on the distributed framework of the proposed P2D3PG algorithm, the computing resources of UEs for training their actors can be saved. Additionally, the replay buff $\Omega$ on the MEC server does not contain any privacy information of UEs.

{\begin{remark}
Note that while there are actor and critic networks at the MEC server side, we only have actor network at the user side. This arrangement is determined by the function of the critic network in the proposed P2D3PG algorithm. More specifically, the critic network is used for guiding the gradient descent of the actor network parameters during the training phase. Since the entire training phase of the proposed scheme is completed at the MEC server in \textbf{Algorithm~\ref{alg_P2D3PG_mec}}, there is no need to deploy the critic network at the user side.
\end{remark}}

\section{NUMERICAL SIMULATIONS AND ANALYSES}
\label{sec4}

In the simulation, we set the number of total files $N = 24$, and the window length of the extractor $H = 10$. For all the local UEs, we assume their cache capacity are equal, denoted as $M_{i}=M_{j}$, $\forall i,j\in\mathcal{I}$, $i\neq j$. In our simulation, the set ${{\cal G}_i}$ of each local {UE}-$i$ is randomly generated. Besides, the transition probability matrix $\mathbf{P}_{i}=\left[P_{g_{l}g_{k}}^{i}\right]_{g_{l},g_{k}=0}^{G_{i}}$ is also generated randomly, where $P_{g_{l}g_{k}}^{i}$ denotes the transition probability from $\alpha_{g_{l}}^{i}$ to $\alpha_{g_{k}}^{i}$. It should be emphasized that the parameter set ${{\cal G}_i}$ and the transition probability matrix $\mathbf{P}_{i}$ are both unknown neither to the MEC server or {UE}-$i$ itself. Adam optimizer \cite{King14} is used to train the parameters $\Theta^{\textrm{C}}$ and $\Theta^{\textrm{A}}$ with the same adaptive learning rate starting from $10^{-4}$.

\begin{figure}[!b]
\centering
\includegraphics[width=0.42\textwidth]{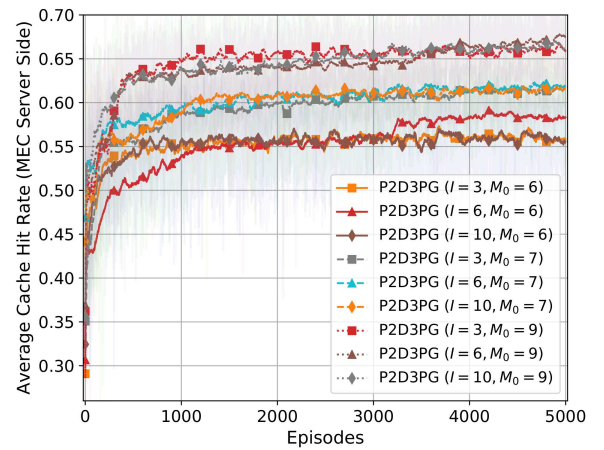}
\caption{Convergence and generalization of the proposed P2D3PG methods in the dynamic edge caching.}
\label{fig7}
\end{figure}

\begin{figure*}[!htbp]
\centering
\subfloat[]{\includegraphics[width=0.9\columnwidth]{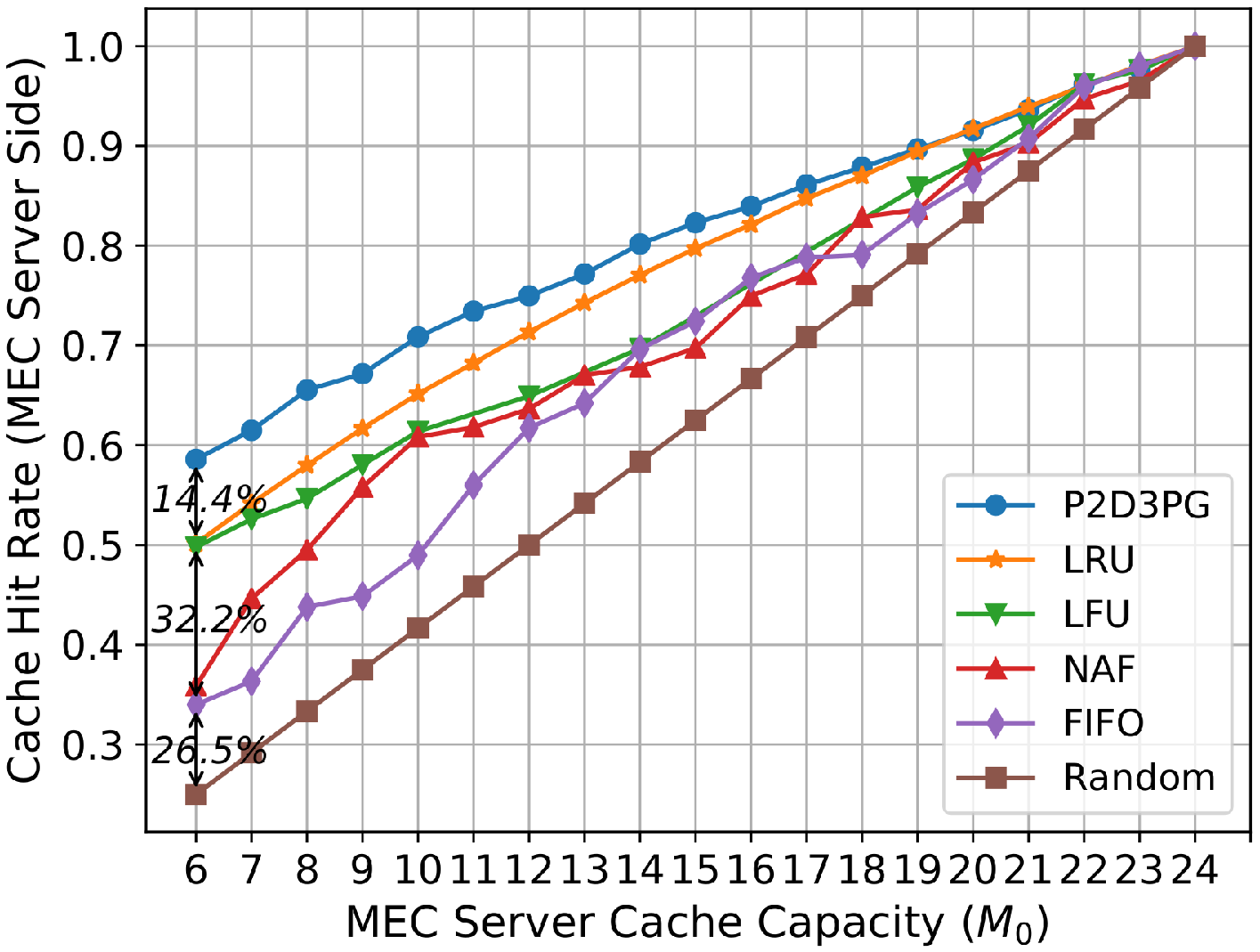}}\hfil
\subfloat[]{\includegraphics[width=0.9\columnwidth]{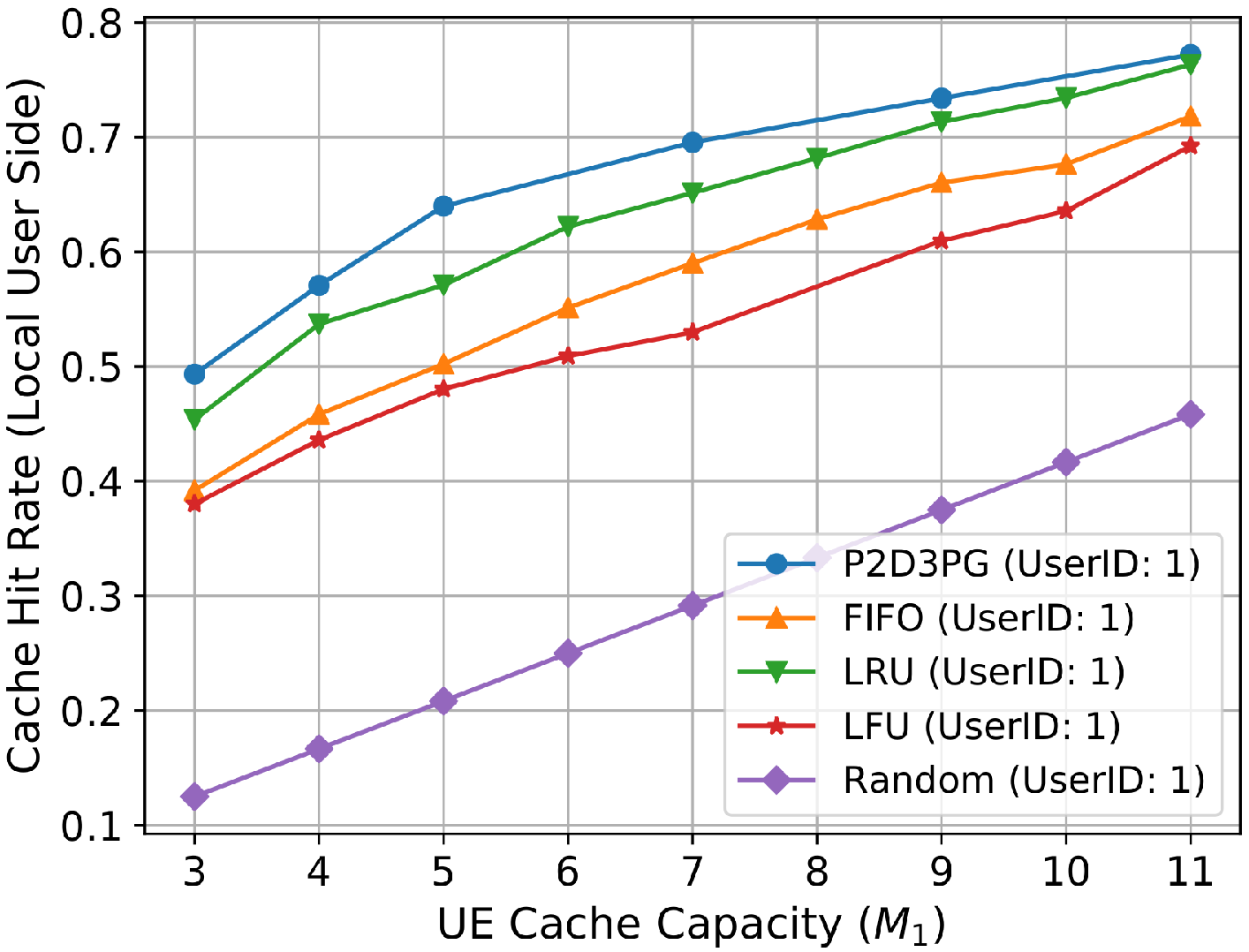}}
\caption{Performance comparison of the proposed P2D3PG methods in terms of cache hit rate with $I=6$. (a) MEC server side. (b) Local user side of UserID~1.}
\label{fig8}
\end{figure*}

Then, we evaluate the performance of the proposed P2D3PG algorithm at the MEC server side and the local user side respectively. There are five baselines for comparison. Three popular methods in distributed EC system including the least recently used (LRU) \cite{Leff96} which discards the least recently used contents, the least frequently used (LFU) \cite{MaG17} which discards the least reference contents in the cache, and the first input first output (FIFO) which discards the initial contents in the FIFO queue. These three methods realize the cache update without privacy preservation. For LRU and LFU, they both need to record the UEs' request information continually in order to count the contents' requested frequency. For FIFO, it needs to maintain a queue of the request information which contains UEs' privacy. Additionally, we also set a normalized advantage functions (NAF) method \cite{Gu16} as another baseline. The NAF algorithm is a deep reinforcement learning algorithm developed based on the deep Q network algorithm \cite{Volo15} and is applicable to the high-dimensional action control problem. For the training of the NAF algorithm, historical request information of UEs must to be collected and stored in the MEC server without any consideration of the privacy preservation. Lastly, in the baseline of random method, the caching policy is randomly formulated and a random action is executed regardless of the current state.

\begin{figure*}[!htbp]
\centering
\subfloat[]{\includegraphics[width=0.9\columnwidth]{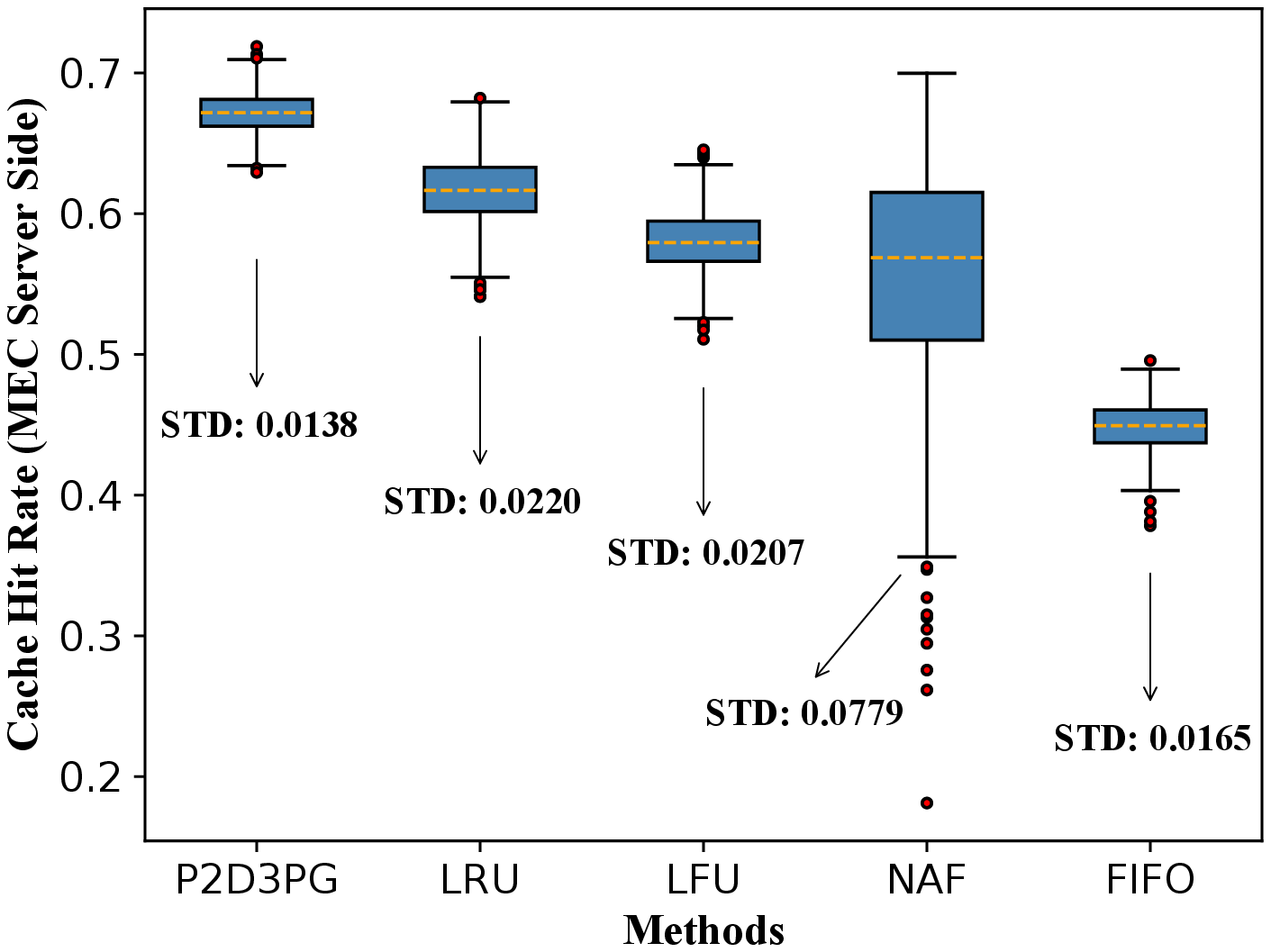}}\hfil
\subfloat[]{\includegraphics[width=0.9\columnwidth]{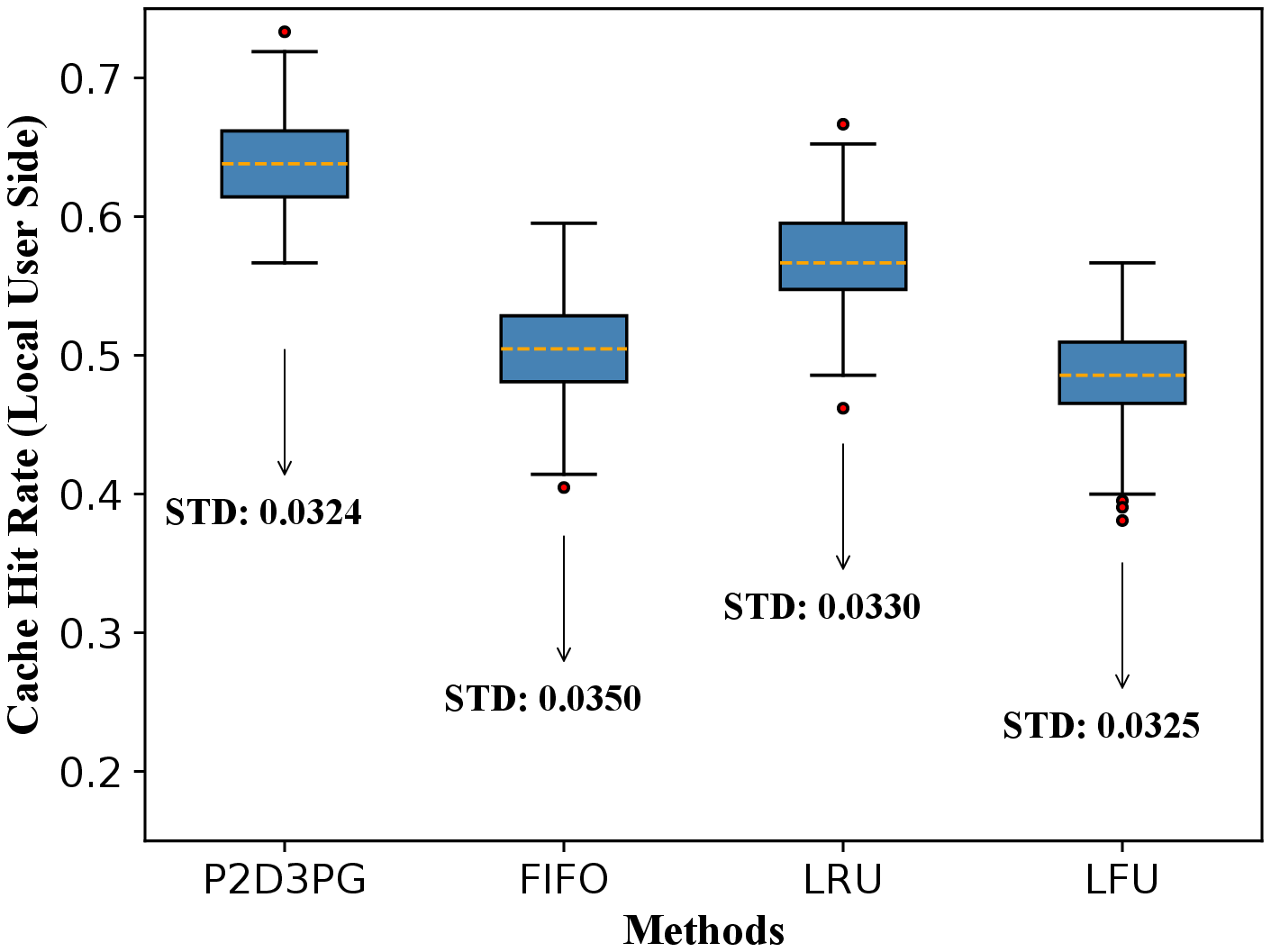}}
\caption{{Performance comparison in terms of standard deviation of the cache hit rate with $I=6$, $H=10$, and $N=24$. (a) MEC server side. ($M_0=9$) (b) Local user side of UserID~1. ($M_1=5$)}}
\label{std}
\end{figure*}

From the perspective of convergent behavior of the proposed P2D3PG, the training processes of the proposed P2D3PG under different $I$ and $M_0$ are illustrated in Fig. \ref{fig7}. We can observe from Fig. \ref{fig7} that the MEC server can achieve a stabilized mean of the cache hit rate around $4000$ episodes under different number of UEs or different cache capacity, which indicates that the agent at the MEC server has acquired the inner knowledge within the global region and the proposed P2D3PG algorithm gradually converges. We also observe from Fig. \ref{fig7} that, the P2D3PG algorithm can achieve basically similar performance when the cache capacity of the MEC server is fixed but the number of UEs within the service coverage is changed. Besides, we can see that the average cache hit rate of the MEC server increases with increasing $M_0$, since the larger cache capacity can cache more effective contents for the UEs.

At the MEC server side, the performance comparison among the proposed P2D3PG algorithm and all the baseline methods versus the cache capacity $M_0$ from 6 units to 24 units are presented in Fig. \ref{fig8}(a). It can be seen from Fig. \ref{fig8}(a) that the proposed P2D3PG algorithm outperforms all the other baseline methods in terms of cache hit rate at the MEC server side while ensuring privacy preservation. When the extreme $M_{0}=N=24$ happens, all the considered methods can reach to $100\%$ cache hit rate. The reason is that the EC optimization aims at utilizing the limited cache resources more effective. When the MEC server can cache all the possible contents of the service, there is no sense to optimize the EC policy and all the requests can always be satisfied. Furthermore, Fig. \ref{fig8}(a) also shows that the advantage of the proposed P2D3PG method becomes more significant as the cache capacity of the MEC server decreases. This implies that the proposed P2D3PG is more competitive with regard to the cache hit rate especially when the cache resource of the MEC server is limited. Specifically, the cache hit rate of the proposed P2D3PG is about $60\%$ when the cache capacity $M_0=6$ is $25\%$ of the total contents' size, which is nearly $14.4\%$ higher than the LRU and LFU methods, $46.6\%$ higher than the NAF and FIFO methods, and almost $73.1\%$ higher than the random baseline methods. Note that, the privacy preservation of the proposed P2D3PG method is another advantage over the baselines. The performance comparisons between the proposed P2D3PG algorithm and all the baseline methods versus the UE cache capacity $M_i$ from $3$ units to $11$ units at the end side are presented in Fig. \ref{fig8}(b). The evaluation of the local UEs is represented by user with UserID 1. We can observe from Fig. \ref{fig8}(b) that the proposed P2D3PG algorithm still outperform all the other baseline methods in terms of cache hit rate at the local user side while realizes the privacy preservation.

\begin{figure*}[!htbp]
\centering
\subfloat[]{\includegraphics[width=0.9\columnwidth]{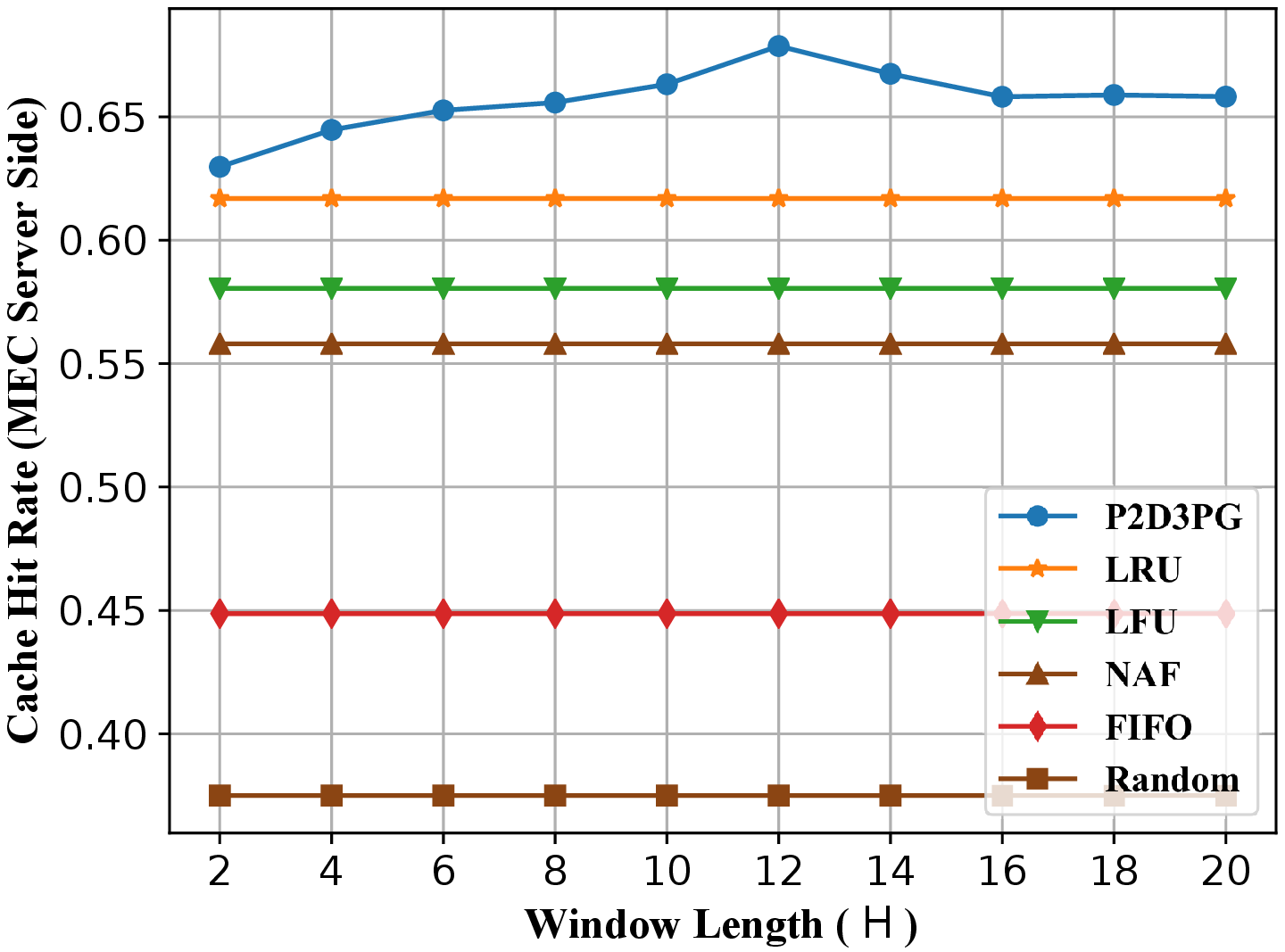}}\hfil
\subfloat[]{\includegraphics[width=0.9\columnwidth]{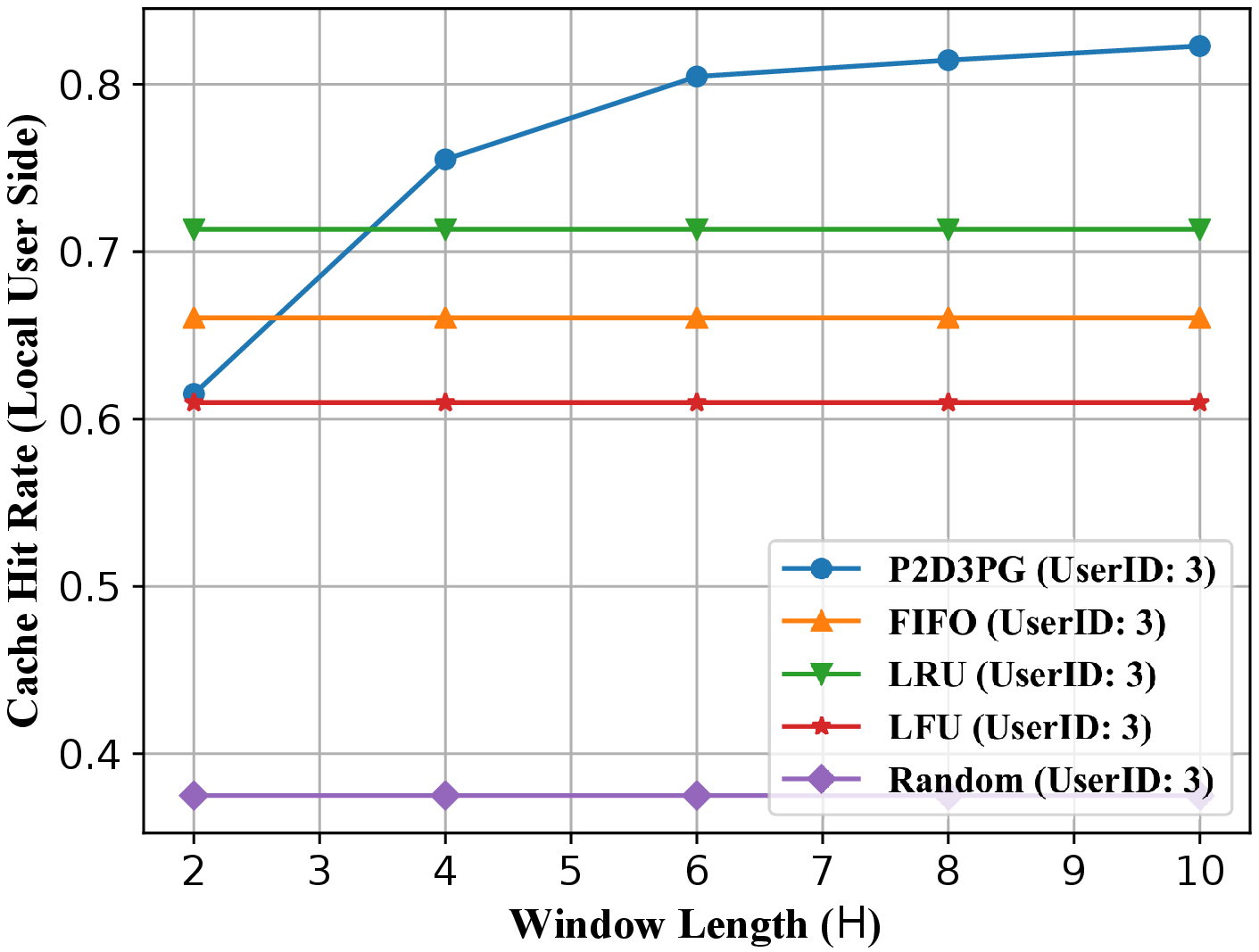}}
\caption{{Impact of the window length $H$ of the extractor on the cache-hit-rate performance with $I=10$ and $N=24$. (a) MEC server side. ($M_0=9$) (b) Local user side of UserID 3. ($M_3=9$)}}
\label{diff_H}
\end{figure*}

\begin{figure*}[!htbp]
\centering
\subfloat[]{\includegraphics[width=0.9\columnwidth]{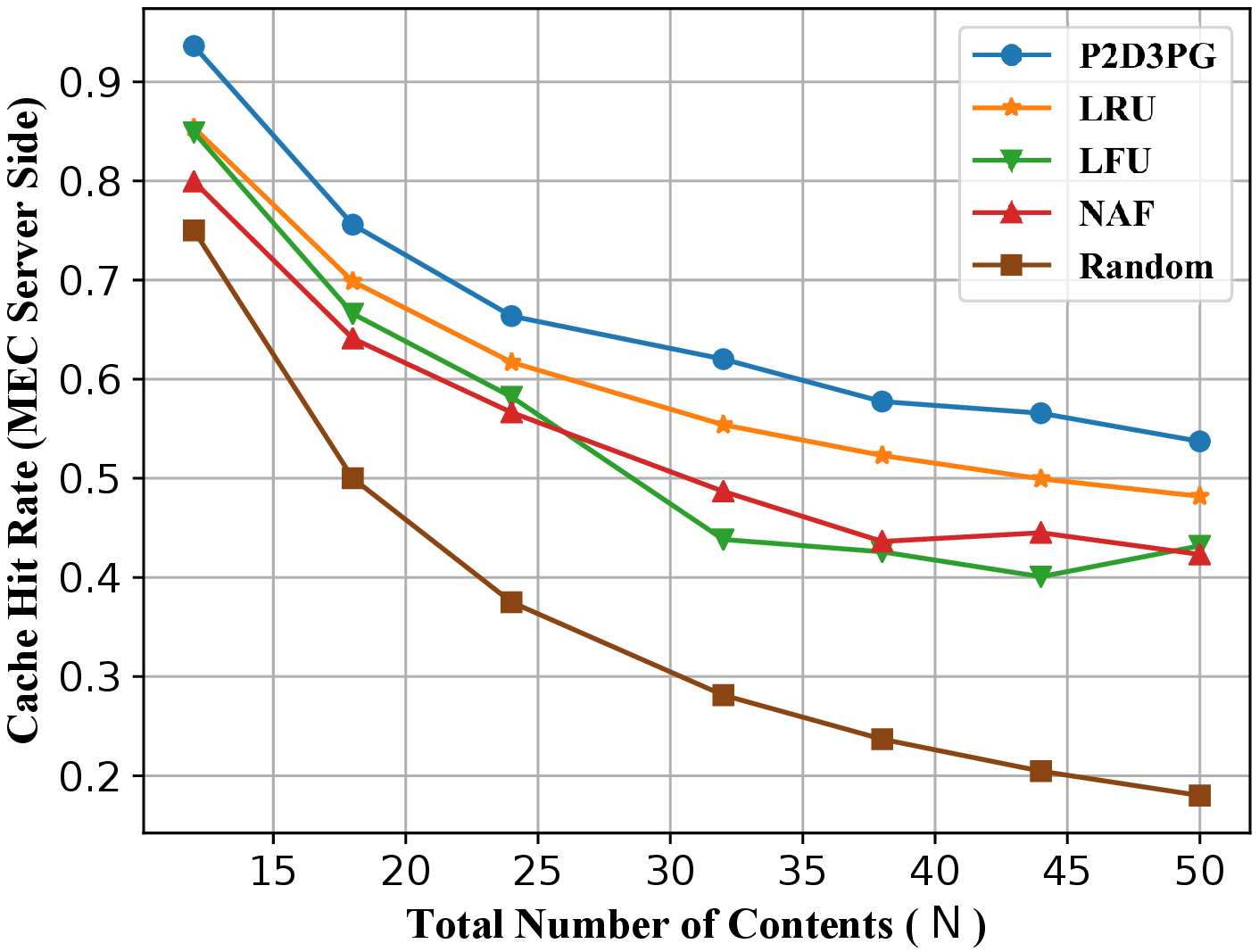}}\hfil
\subfloat[]{\includegraphics[width=0.9\columnwidth]{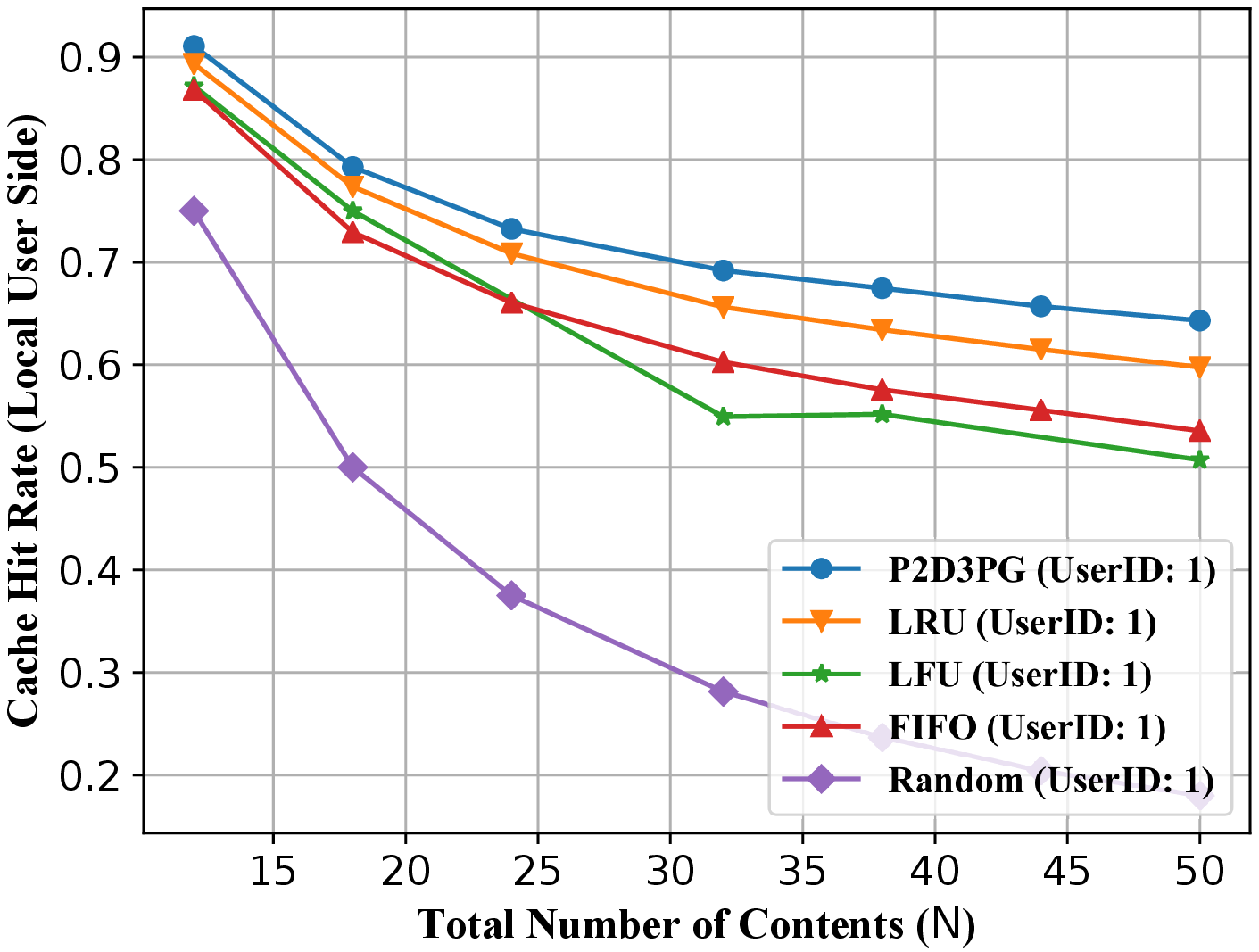}}
\caption{{Impact of the number of total contents $N$ on the cache-hit-rate performance with $I=10$ and $H=10$. (a) MEC server side. ($M_0=9$) (b) Local user side of UserID 1. ($M_1=9$)}}
\label{diff_N}
\end{figure*}

{As described earlier, in the proposed scheme, we formulate an optimization problem to predict the upcoming files which are going to be requested by users. Henceforth, the cache hit rate is improved by averaging over time. While the goal is to maximize the average cache hit rate, it is also meaningful to examine the standard deviation (SD) of cache hit rate at both the MEC and the local user sides. At the MEC server side, we test all the methods within a period of $1024$ continuous time slots with $I=6$ and $M_0=9$. We record all the testing results of the cache hit rate $H_{0}\left(t\right)$ to draw Fig. \ref{std}(a) and calculate their SDs. We observe from Fig. \ref{std}(a) that, at the MEC server side, the proposed P2D3PG algorithm can achieve the lowest SD at $0.0138$ while yielding the highest cache hit rate compared to the baseline methods. Regarding the local user side, we test all the methods on UE-$1$ during $512$ continuous time slots with $M_1=5$. Then, we also visualize the results of $H_{i}^{\textrm{savg}}\left(t\right)$, which is given in Fig.~\ref{std}(b). Although at the local user side all the compared methods obtain very close SDs, P2D3PG is still superior as it achieves the highest cache hit rate as well as preserves users' privacy.}

{We further explore the effect of different window length $H$ on the cache hit rate. As $H$ is a peculiar parameter of P2D3PG, the curves of the baseline methods are presented to examine if there exists a certain set of model parameters such that the conventional approaches works better or similar to the proposed P2D3PG. We observe from Fig.~\ref{diff_H}(a) that, with the increase of $H$, the cache hit rate at the MEC server side first rises until reaching a certain point and then gradually declines to a steady level. We believe that this is because with an excessively long window length, the algorithm will observe too much redundant information from the historical requests; while if the window is too short, the algorithm can hardly observe sufficient information from the historical requests. From Fig.~\ref{diff_H}(b), we can see that the conventional approaches achieve better cache hit rate than the proposed P2D3PG when $H\leq3$, which also confirms effect of excessively short window length. To recap, unreasonable window length can affect the feature extraction performance of the predictive models and further reduces the prediction accuracy of the popularities. This in turn leads to a decrease of the cache hit rate.}

\begin{figure*}[!htbp]
\centering
\subfloat[]{\includegraphics[width=0.9\columnwidth]{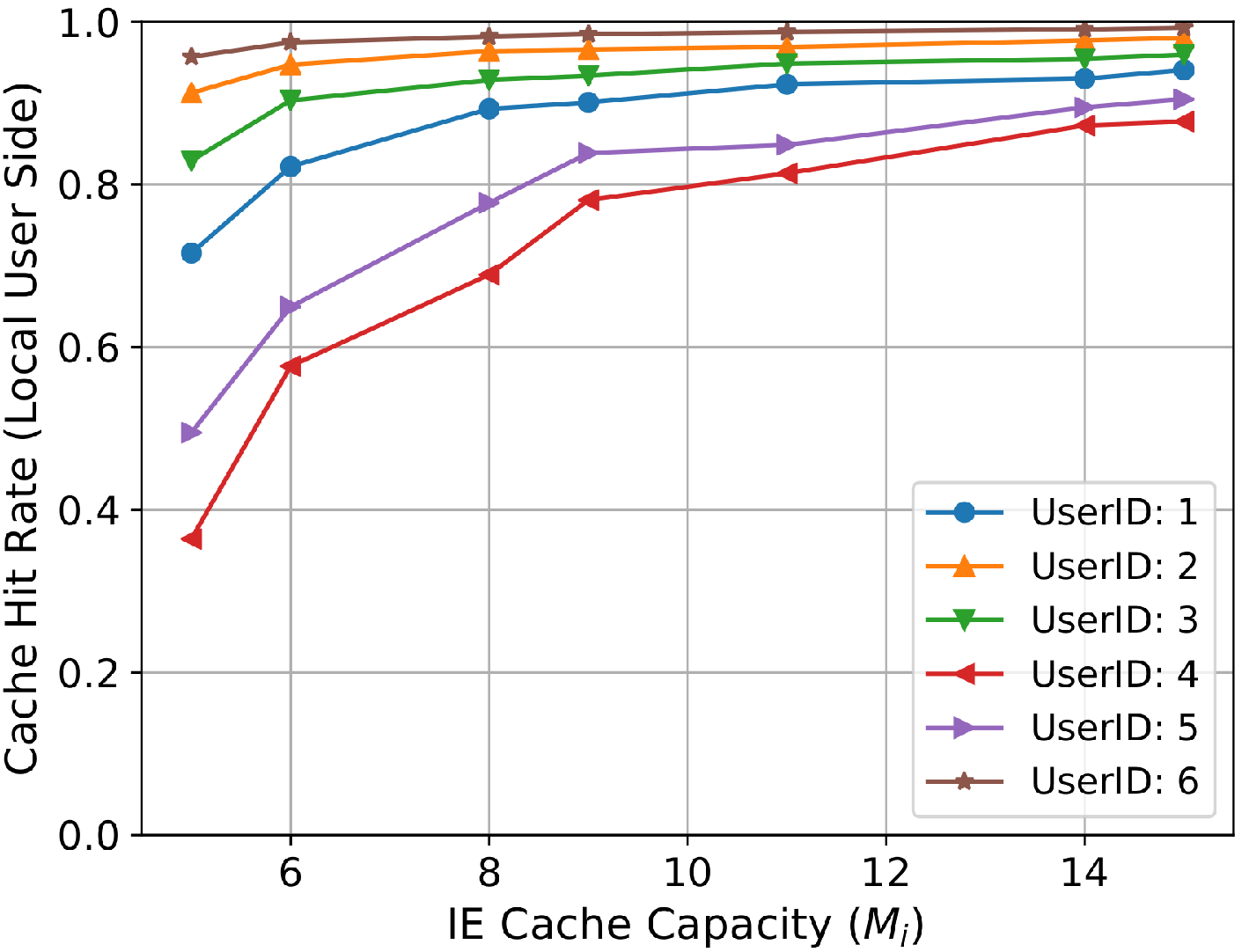}}\hfil
\subfloat[]{\includegraphics[width=0.9\columnwidth]{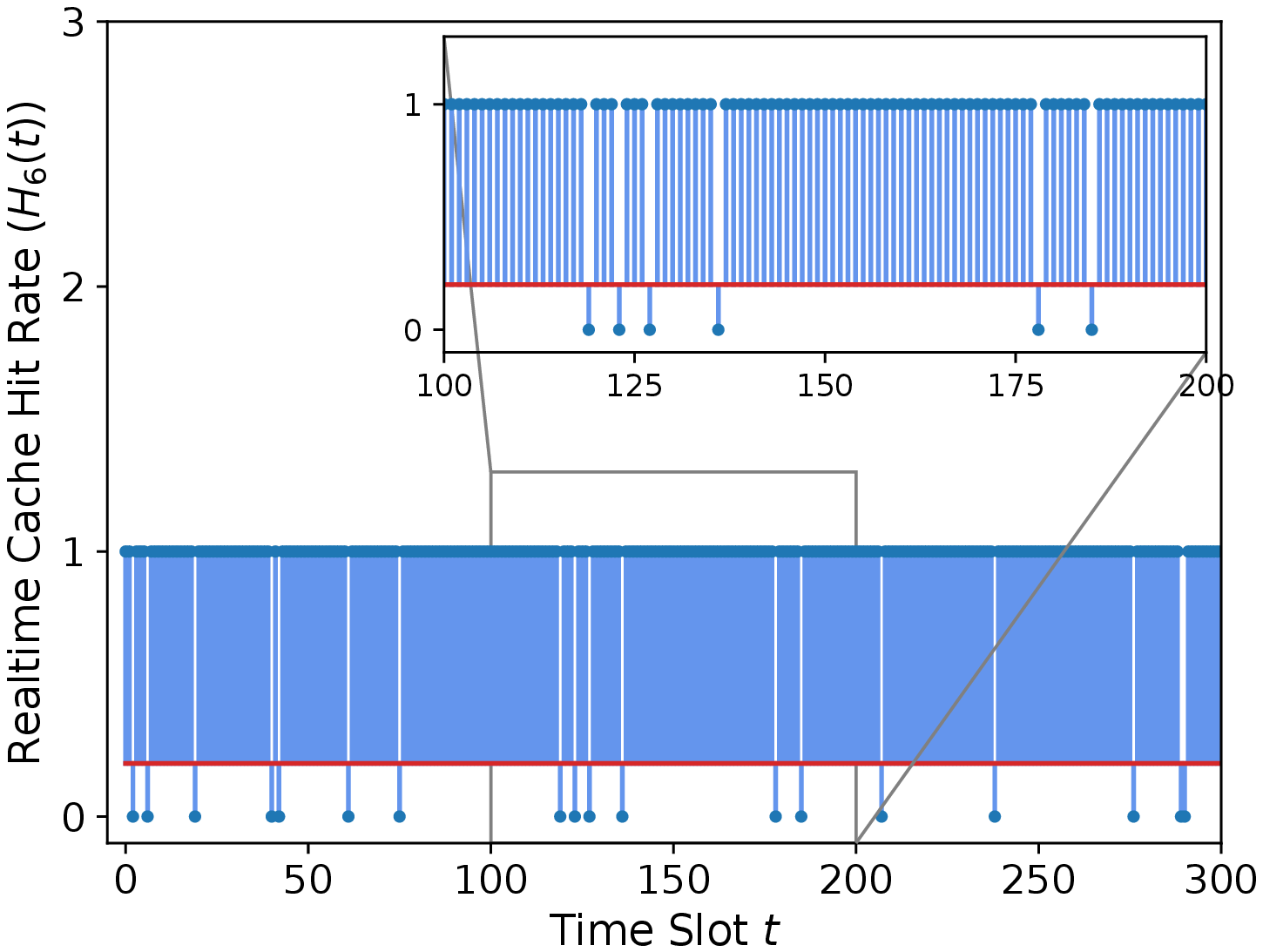}}
\caption{Performance evaluation of the proposed P2D3PG method at the local user side. (a) Average cache hit rate of all the UEs. (b) Realtime cache hit rate of UserID 6.}
\label{fig9}
\end{figure*}

{Likewise, we provide the cache-hit-rate performance comparison with respect to the number of total contents $N$ from $12$ to $50$. Fig.~\ref{diff_N} indicates that the proposed P2D3PG outperforms all the baseline methods at both the MEC server and the local user sides, while their individual cache hit rates drop with an increasing $N$. We also note from Fig.~\ref{diff_N} that, the advantage of the proposed P2D3PG method becomes less pronounced as the number of total contents decreases, which reconfirms the our speculation from Fig.~\ref{fig8}.}

Fig. \ref{fig9}(a) illustrates the performance evaluation of the proposed P2D3PG method at the end side under $I=6$. In particularly, we picked UEs with user identity document (UserID) 1 through 6 in the previous subsection. It can be found that the cache hit rate of each {UE} increases with the cache capacity. In addition, we observe that there are differences in the cache hit rate of different UEs, which results from the independent but not identically distributed behaviors of UEs. For UEs whose variations of popularities are more complicated, the challenges of the popularity predictions by the URFL method are heavier. Thus the prediction accuracies of UEs are different, which results in the different cache hit rates among UEs. We further test the performance of the proposed P2D3PG algorithm with respect to realtime cache hit rate at the end side, which is presented in Fig. \ref{fig9}(b). In Fig. \ref{fig9}(b), the {UE} with UserID of $6$ is taken as an example, the cache capacity $M_6$ is set at 5 units which is $20.8\%$ of the total contents size, and the time window length of the observation is set as $300$ time slots. According to the equation \eqref{e5}, $H_{0}\left(t\right)=0$ when the requested content of {UE}~6 at time $t$ is absent at its current cache. Otherwise, $H_{0}\left(t\right)=1$ when the requested content of {UE}~6 at time $t$ is cached in its local {UE} in advance. On this basis, we find from Fig. \ref{fig9}(b) that the realtime cache hit rate can stay at $100\%$ for the most time slots, which implies that the requested content at most time can be directly satisfied by its local cache. Fig. \ref{fig9}(b) again confirms the superiority of the proposed P2D3PG algorithm on dynamic EC while preserve UEs' privacy.

\section{Conclusion}\label{sec6}

In this paper, the problem of distributed EC hit rate maximization in an MEC-enabled wireless communication system is formulated under time-varying and unobservable content popularities. To address the challenges of distributed problem under the constraints of privacy preservation, a P2D3PG algorithm is proposed to maximize the EC hit rates in the MEC system. The superior performance of the proposed methods compared to the baseline methods are confirmed by numerical simulations. Our future work will concentrate on more complicated scenarios such as heterogeneous multiple MEC nodes as well as further addressing the challenges brought from non-IID user behaviors.

\ifCLASSOPTIONcaptionsoff
  \newpage
\fi

\balance


\begin{thebibliography}{99}


\bibitem{GlobeCom21}
{C.\ Zheng, S.\ Liu, Y.\ Huang, and T.~Q.~S.\ Quek, ``Privacy-preserving federated reinforcement learning for popularity-assisted edge caching,'' in {\em Proc. 40$^{th}$ IEEE Global Commun. Conf. (GLOBECOM'21): Mach. Learn. Commun. Symp.}, Madrid, Spain, Dec. 2021, pp.\ 1--6.}


%
\bibitem{HuF20}
F.\ Hu, Y.\ Deng, W.\ Saad, et al., "Cellular-connected wireless virtual reality: Requirements, challenges, and solutions,'' {\em IEEE Commun. Mag.}, vol.\ 58, no.\ 5, pp.\ 105--111, May 2020.

\bibitem{Duan20}
W.\ Duan, J.\ Gu, M.\ Wen, et al., ''Emerging technologies for 5G-IoV networks: Applications, trends and opportunities,'' {\em IEEE Network}, vol.\ 34, no.\ 5, pp.\ 283--289, Oct.\ 2020.

\bibitem{Abde19}
A.~A.\ Abdellatif, A.\ Mohamed, C.~F.\ Chiasserini, et al., ``Edge computing for smart health: Context-aware approaches, opportunities, and challenges,'' {\em IEEE Network}, vol.\ 33, no.\ 3, pp.\ 196--203, Jun.\ 2019.

\bibitem{Fara20}
G.\ Faraci, C.\ Grasso, and G.\ Schembra, ''Design of a 5G network slice extension with MEC UAVs managed with reinforcement learning,'' {\em IEEE J. Sel. Areas Commun.}, vol.\ 38, no.\ 10, pp.\ 2356--2371, Oct.\ 2020.

\bibitem{DuJ20}
J.\ Du, F.~R.\ Yu, G.\ Lu, et al., ``MEC-assisted immersive VR video streaming over terahertz wireless networks: A deep reinforcement learning approach," {\em IEEE Internet Things J.}, vol.\ 7, no.\ 10, pp.\ 9517--9529, Oct.\ 2020.

\bibitem{Xiong20}
X.\ Xiong, K.\ Zheng, L.\ Lei, and L.\ Hou, ``Resource allocation based on deep reinforcement learning in IoT edge computing,'' {\em IEEE J. Sel. Areas Commun.}, vol.\ 38, no.\ 6, pp.\ 1133--1146, Jun.\ 2020.

\bibitem{Miao18}
M.\ Du, K.\ Wang, Y.\ Chen, et al., ``Big data privacy preserving in multi-access eEdge computing for heterogeneous internet of things,'' {\em IEEE Commun. Mag.}, vol.\ 56, no.\ 8, pp.\ 62--67, Aug.\ 2018.

\bibitem{Zhao20}
Z.\ Zhao, R.\ Zhao, J.\ Xia, et al., ``A novel framework of three-hierarchical offloading optimization for MEC in industrial IoT networks,'' {\em IEEE Trans. Ind. Inf.}, vol.\ 16, no.\ 8, pp.\ 5424--5434, Aug.\ 2020.

\bibitem{Wang20}
X.\ Wang, C.\ Wang, X.\ Li, et al., ''Federated deep reinforcement learning for internet of things with decentralized cooperative edge caching," {\em IEEE Internet Things J.}, vol.\ 7, no.\ 10, pp.\ 9441--9455, Oct.\ 2020.

%
%
%


%

\bibitem{cui16}
Y.\ Cui, D.\ Jiang, and Y.\ Wu, ``Analysis and optimization of caching and multicasting in large-scale cache-enabled wireless networks,'' {\em IEEE Trans. Wireless Commun.}, vol.\ 15, no.\ 7, pp.\ 5101--5112, Jul.\ 2016.

\bibitem{Niko16}
S.\ Nikolaou, R.~V.\ Renesse, and N.\ Schiper, ``Proactive cache placement on cooperative client caches for online social networks,'' {\em IEEE Trans. Parallel Distrib. Syst.}, vol.\ 27, no.\ 4, pp.\ 1174--1186, Apr.\ 2016.

\bibitem{LiQ20}
Q.\ Li, Y.\ Zhang, Y.\ Li, et al., ``Capacity-aware edge caching in fog computing networks,'' {\em IEEE Trans. Veh. Technol.}, vol.\ 69, no.\ 8, pp.\ 9244--9248, Aug.\ 2020.

\bibitem{Jiang19}
Y.\ Jiang, M.\ Ma, M.\ Bennis, et al., ``User preference learning-based edge caching for fog radio access network,'' {\em IEEE Trans. Commun.}, vol.\ 67, no.\ 2, pp.\ 1268--1283, Feb.\ 2019.

\bibitem{Azimi18}
S.~M.\ Azimi, O.\ Simeone, A.\ Sengupta, and R.\ Tandon, ``Online edge caching and wireless delivery in fog-aided networks with dynamic content popularity,'' {\em IEEE J. Sel. Areas Commun.}, vol.\ 36, no.\ 6, pp.\ 1189--1202, Jun.\ 2018.

\bibitem{Liu20}
J.\ Liu, D.\ Li, and Y.\ Xu, ``Collaborative online edge caching with bayesian clustering in wireless networks,'' {\em IEEE Internet Things J.}, vol.\ 7, no.\ 2, pp.\ 1548--1560, Feb.\ 2020.


%




\bibitem{DaiY20}
Y.\ Dai, D.\ Xu, K.\ Zhang, et al., ``Deep reinforcement learning and permissioned blockchain for content caching in vehicular edge computing and networks," {\em IEEE Trans. Veh. Technol.}, vol.\ 69, no.\ 4, pp.\ 4312--4324, Apr.\ 2020.

\bibitem{XuQ19}
Q.\ Xu, Z.\ Su, Q.\ Zheng, et al., ``Game theoretical secure caching scheme in multihoming edge computing-enabled heterogeneous networks,'' {\em IEEE Internet Things J.}, vol.\ 6, no.\ 3, pp.\ 4536--4546, Jun.\ 2019.

\bibitem{Xiao18}
L.\ Xiao, X.\ Wan, C.\ Dai, X.\ Du, X.\ Chen, and M.\ Guizani, ``Security in mobile edge caching with reinforcement learning,'' {\em IEEE Wireless Commun.}, vol.\ 25, no.\ 3, pp.\ 116--122, Jun.\ 2018.


\bibitem{YuZ20}
Z.\ Yu, J.\ Hu, G.\ Min, et al., ``Mobility-aware proactive edge caching for connected vehicles using federated learning,'' {\em EEE Trans. Intell. Transp. Syst.}, to be published, doi: 10.1109/TITS.2020.3017474.
%

\bibitem{Sade17}
A.\ Sadeghi, F.\ Sheikholeslami, and G.~B.\ Giannakis, ``Optimal and scalable caching for 5G using reinforcement learning of space-time popularities,'' {\em IEEE J. Sel. Top. Signal Process.}, vol.\ 12, no.\ 1, pp.\ 180--190, Feb.\ 2018.

%

\bibitem{Zheng20b}
C.\ Zheng, S.\ Liu, Y.\ Huang, and L.\ Yang, ``MEC-enabled wireless VR video service: A learning-based mixed strategy for energy-latency tradeoff,'' in {\em Proc. 18$^{th}$ IEEE Wireless Commun. Netw. Conf. (WCNC'20)}, Seoul, South Korea, Apr. 2020, pp.\ 1--6.

\bibitem{Zheng20}
C.\ Zheng, S.\ Liu, Y.\ Huang, and L.\ Yang, ``Hybrid policy learning for energy-latency tradeoff in MEC-assisted VR video service,'' {\em IEEE Trans. Veh. Technol.}, vol.\ 70, no.\ 9, pp.\ 9006--9021, Sept.\ 2021.

\bibitem{Sutt98}
R.\ Sutton and A.\ Barto, {\em Reinforcement Learning: An Introduction,} Cambridge, MA, USA: MIT press, 1998.

\bibitem{King14}
D.\ Kingma and J.\ Ba, ``Adam: A method for stochastic optimization,'' in {\em Proc. 3$^{rd}$ Int. Conf. Learn. Represent. (ICLR'15)}, San Diego, CA, USA, May 2015.






\bibitem{Leff96}
A.\ Leff, J.~L.\ Wolf, and P.~S.\ Yu, ``Efficient LRU-based buffering in a LAN remote caching architecture,'' {\em IEEE Trans. Parallel Distrib. Syst.}, vol.\ 7, no.\ 2, pp.\ 191--206, Feb.\ 1996.

\bibitem{MaG17}
G.\ Ma, Z.\ Wang, M.\ Zhang, et al., ``Understanding performance of edge content caching for mobile video streaming,'' {\em IEEE J. Sel. Areas Commun.}, vol.\ 35, no.\ 5, pp.\ 1076--1089, May 2017.

\bibitem{Gu16}
S.\ Gu, T.\ Lillicrap, I.\ Sutskever, and S.\ Levine, ``Continuous deep Q-learning with model-based acceleration,'' in {\em Proc. 33$^{rd}$ Int. Conf. Mach. Learn. (ICML'16)}, New York, NY, USA, June 2016, pp.\ 2829--2838.

\bibitem{Volo15}
M.\ Volodymyr, K.\ Koray, S.\ David, et al., ``Human-level control through deep reinforcement learning,'' {\em Nature}, vol.\ 518, no.\ 7540, pp.\ 529--533, Feb.\ 2015.

\bibitem{Kair20}
P.\ Kairouz, H.~B.\ McMahan, B.\ Avent, et al., ``Advances and open problems in federated learning,'' {\em arXiv preprint arXiv:1912.04977}, 2019.

\end{thebibliography}
\end{document}